\documentclass[times,twocolumn,final]{elsarticle}
\usepackage[table]{xcolor}
\usepackage{cag}
\usepackage{amssymb}
\usepackage{graphicx}
\usepackage{arydshln} 
\usepackage{multirow}
\usepackage{booktabs}
\usepackage{subcaption}
\usepackage{hyperref}
\hypersetup{
    colorlinks=true,
    linkcolor=blue,
    filecolor=magenta,      
    urlcolor=cyan,
}

\usepackage{amsmath}
\usepackage{amsfonts}
\usepackage{orcidlink}
\usepackage[utf8]{inputenc}
\usepackage[T1]{fontenc}
\usepackage{colortbl}
\definecolor{lightgray}{gray}{0.9}

\newcommand{\highlight}[1]{\textcolor{black}{#1}}

\journal{Computers \& Graphics}

\begin{document}

\begin{frontmatter}

\title{SHREC 2025: Retrieval of Optimal Objects for Multi-modal Enhanced Language and Spatial Assistance (ROOMELSA)}

\author[1,3]{Trong-Thuan Nguyen\orcidlink{0000-0001-7729-2927}}
\author[1,3]{Viet-Tham Huynh\orcidlink{0000-0002-8537-1331}}
\author[1,3]{Quang-Thuc Nguyen\orcidlink{0000-0003-2523-8851}}
\author[1,3]{Hoang-Phuc Nguyen\orcidlink{0009-0000-2946-7956}}
\author[2,3]{Long Le Bao\orcidlink{0009-0004-6079-6116}}
\author[2,3]{Thai Hoang Minh\orcidlink{0009-0008-0577-1484}}
\author[2,3]{Minh Nguyen Anh\orcidlink{0009-0009-2228-0764}}
\author[2,3]{Thang Nguyen Tien\orcidlink{0009-0008-9139-9884}}
\author[2,3]{Phat Nguyen Thuan\orcidlink{0009-0005-4672-2185}}
\author[2,3]{Huy Nguyen Phong\orcidlink{0009-0005-8396-9152}}
\author[2,3]{Bao Huynh Thai\orcidlink{0009-0001-7489-1059}}
\author[2,3]{Vinh-Tiep Nguyen\orcidlink{0000-0003-4260-7874}}
\author[2,3]{Duc-Vu Nguyen\orcidlink{0000-0003-0072-2524}}
\author[1,3]{Phu-Hoa Pham\orcidlink{0009-0001-5471-2578}}
\author[1,3]{Minh-Huy Le-Hoang\orcidlink{0009-0005-6064-5585}}
\author[1,3]{Nguyen-Khang Le\orcidlink{0009-0004-2644-7828}}
\author[2,3]{Minh-Chinh Nguyen\orcidlink{0009-0004-3212-104X}}
\author[2,3]{Minh-Quan Ho\orcidlink{0009-0005-7531-8894}}
\author[2,3]{Ngoc-Long Tran\orcidlink{0009-0007-3119-7634}}
\author[2,3]{Hien-Long Le-Hoang\orcidlink{0009-0002-5593-8648}}
\author[2,3]{Man-Khoi Tran\orcidlink{0009-0003-7246-7101}}
\author[2,3]{Anh-Duong Tran\orcidlink{0009-0002-0393-5779}}
\author[1,3]{Kim Nguyen\orcidlink{0009-0001-4609-6381}}
\author[2,3]{Quan Nguyen Hung\orcidlink{0009-0001-7294-5543}}
\author[2,3]{Dat Phan Thanh\orcidlink{0009-0001-3694-4623}}
\author[2,3]{Hoang Tran Van\orcidlink{0009-0007-9029-4442}}
\author[2,3]{Tien Huynh Viet\orcidlink{0009-0004-4356-2211}}
\author[2,3]{Nhan Nguyen Viet Thien\orcidlink{0009-0000-7068-6798}}
\author[1,3]{Dinh-Khoi Vo\orcidlink{0000-0001-8831-8846}}
\author[1,3]{Van-Loc Nguyen\orcidlink{0000-0001-9351-3750}}
\author[1,3]{Trung-Nghia Le\orcidlink{0000-0002-7363-2610}} 

\author[4]{Tam V. Nguyen\orcidlink{0000-0003-0236-7992}}
\author[1,3]{Minh-Triet Tran\orcidlink{0000-0003-3046-3041}\corref{cor1}} \ead{tmtriet@fit.hcmus.edu.vn}

\affiliation[1]{organization={University of Science, VNU-HCM}, city={Ho Chi Minh City}, country={Vietnam}}
\affiliation[2]{organization={University of Information Technology, VNU-HCM}, city={Ho Chi Minh City}, country={Vietnam}}
\affiliation[3]{organization={Vietnam National University}, city={Ho Chi Minh City}, country={Vietnam}}
\affiliation[4]{organization={University of Dayton}, city={Ohio}, country={U.S.}}

\cortext[cor1]{Corresponding author}

\received{\textit{May 14, 2025}}
\accepted{\textit{31 July, 2025}}

\begin{abstract}

Recent 3D retrieval systems are typically designed for simple, controlled scenarios, such as identifying an object from a cropped image or a brief description. However, real-world scenarios are more complex, often requiring the recognition of an object in a cluttered scene based on a vague, free-form description. To this end, we present ROOMELSA, a new benchmark designed to evaluate a system’s ability to interpret natural language. Specifically, ROOMELSA attends to a specific region within a panoramic room image and accurately retrieves the corresponding 3D model from a large database. In addition, ROOMELSA includes over 1,600 apartment scenes, nearly 5,200 rooms, and more than 44,000 targeted queries. Empirically, while coarse object retrieval is largely solved, only one top-performing model consistently ranked the correct match first across nearly all test cases. Notably, a lightweight CLIP-based model also performed well, although it struggled with subtle variations in materials, part structures, and contextual cues, resulting in occasional errors. These findings highlight the importance of tightly integrating visual and language understanding. By bridging the gap between scene-level grounding and fine-grained 3D retrieval, ROOMELSA establishes a new benchmark for advancing robust, real-world 3D recognition systems.

\end{abstract}





\end{frontmatter}


\section{Introduction}\label{sec:introduction}
\begin{figure*}[!t]
    \centering
    \includegraphics[width=\linewidth]{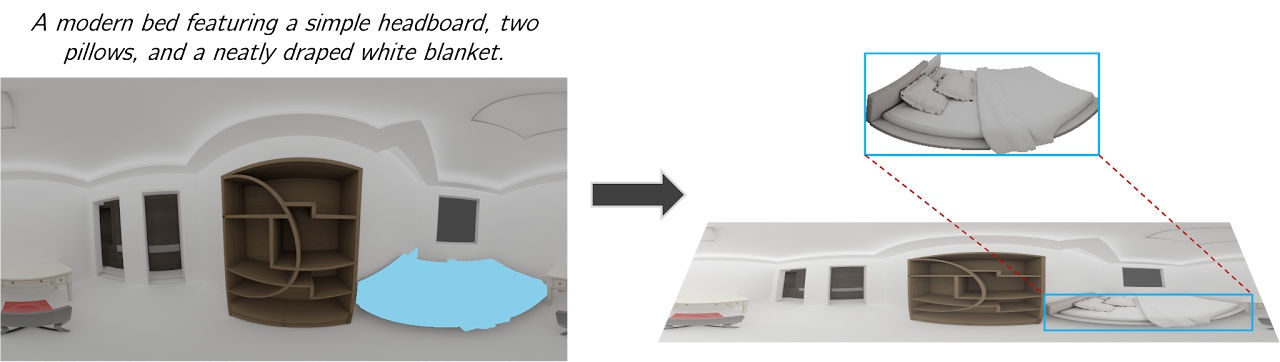}
    \caption{Our ROOMELSA challenge. From left to right: a cluttered 3-D apartment rendered as an equirectangular panorama; a binary mask in \textcolor{cyan}{cyan} that highlights the query region; and the gallery-retrieval stage, where the system must rank the single CAD mesh that matches the masked object above all other candidates.}
    \label{fig:task}
\end{figure*}

The recent proliferation of large-scale indoor scans and publicly available 3D repositories has significantly advanced scene-level understanding, moving beyond the analysis of isolated objects. Tasks such as 3D semantic segmentation, object detection, and visual grounding have benefited substantially from richly annotated datasets, including ScanNet~\cite{dai2017scannet}, Matterport3D~\cite{chang2017matterport3d}, and 3D-Front~\cite{fu20213d}. 
However, the ability to retrieve a specific 3D object from a cluttered scene based on a natural language description and a spatial mask remains largely underexplored. This functionality is essential for real-world applications ranging from context-aware asset replacement in game engines and augmented reality interior design to robotic manipulation, where a gripper must accurately identify and retrieve a target object amidst similar distractors on a crowded surface.

Previous benchmarks addressing language–vision alignment in 3D environments, such as ReferIt3D~\cite{achlioptas2020referit3d}, ScanRefer~\cite{chen2020scanrefer}, and ViGiL3D~\cite{wang2025vigil3d}, typically formulate the task as object localization. Given a natural language description, the objective is to predict a bounding box or point-cloud cluster corresponding to the referred object within a 3D scan. In contrast, ROOMELSA (Retrieval of Optimal Objects for Multi-modal Enhanced Language and Spatial Assistance) introduces a reversed formulation. Instead of inferring spatial location from language, ROOMELSA assumes a predefined spatial region, specified by a mask, and requires retrieving the 3D model from a large-scale object catalog that best matches the semantics and geometry of the masked region. This reformulation introduces three principal challenges, as illustrated in Fig.~\ref{fig:task}. First, the task presents a significant cross-domain appearance gap, as RGB-D renderings of real-world, cluttered scenes differ markedly from the clean, studio-lit images used to depict catalog exemplars. Second, it demands fine-grained visual discrimination. Especially, many scenes contain multiple instances from the same object category, such as dining chairs, and accurate retrieval hinges on the system’s ability to discern subtle variations in shape, material, and style that are often only implicitly described in language. Third, effective retrieval requires robust spatial reasoning, as referring expressions often contain relational cues (e.g., ``the mug closest to the sink'' or ``the lamp behind the sofa''), which demand a holistic understanding of both the masked region and the broader spatial configuration of the RGB-D rendered scene.

To foster progress in this emerging direction, we introduce the ROOMELSA Challenge as part of SHREC 2025~\footnote{\url{https://www.shrec.net/}} and the first benchmark dataset explicitly designed for this task. The dataset comprises 1,622 apartment-level scenes, encompassing 5,197 individual rooms, and includes a total of 44,445 query pairs (mask, text). The data is divided into public and private test splits. For each masked query, participants were asked to submit a ranked list of ten candidate catalog item IDs, with performance evaluated using the Mean Reciprocal Rank (MRR). Unlike previous challenges centered on generic retrieval or phrase grounding~\cite{Hameed-SHREC2018, Hameed-SHREC2019, Li-SHREC2019, Li-SHREC2020, Feng-SHREC2022, Qin-SHREC2022, le2023sketchanimar, le2023textanimar}, our ROOMELSA challenge introduces three distinctive features. It conditions language queries on predefined spatial masks, includes a significantly broader taxonomy of furniture and décor styles, and supports a zero-shot public test phase to encourage scalable, generalizable solutions without extensive fine-tuning. The 2025 edition attracted 18 teams from across the globe, who explored various algorithmic strategies, including frozen vision–language models, depth-aware point-cloud alignment, multi-view CLIP indexing, mask-guided inpainting, and hybrid voting mechanisms. The top-performing method achieved an MRR of 0.97, setting a benchmark for ROOMELSA and future work.

\noindent\textbf{Contributions.} This paper makes three main contributions:
\begin{itemize}
    \item In the ROOMELSA challenge, we formulate 3D object retrieval as a \emph{mask-conditioned, language-driven} grounding task in 3D environments, unifying spatial reference and natural language understanding within a unified task.
    
    \item We introduce ROOMELSA\footnote{The ROOMELSA dataset is released at \\ \url{https://aichallenge.hcmus.edu.vn/shrec-2025/smart3droom}}, a novel benchmark dataset consisting of 1,622 apartment-scale scenes and 5,197 individual rooms, with 44,445 (mask, text) query pairs.
    
    \item We analyze the top-five challenge entries, revealing how scene-level inference, depth-aware reconstruction, and adaptive cross-modal fusion affect Mean Reciprocal Rank.

\end{itemize}

The remainder of this paper is organized as follows. Sec.~\ref{sec:related_work} reviews related work on language-grounded object retrieval in 3D, referring segmentation, and multimodal foundation models for 3D understanding. Sec.~\ref{sec:discuss} discusses the limitations of previous challenges and highlights the advantages of the proposed ROOMELSA challenge. Sec.~\ref{sec:dataset} introduces the ROOMELSA dataset, defines the task, and describes the evaluation metrics. Sec.~\ref{sec:participants} provides an overview of participant statistics and summarizes the submitted approaches. Sec.~\ref{sec:methods} presents the methods proposed by the top five participating teams. Sec.~\ref{sec:results} reports both quantitative and qualitative results, followed by a detailed analysis. Finally, Sec.~\ref{sec:conclusion} concludes the paper and outlines future research directions enabled by the ROOMELSA benchmark.

\begin{figure*}[!t]
    \centering
    \includegraphics[width=\linewidth]{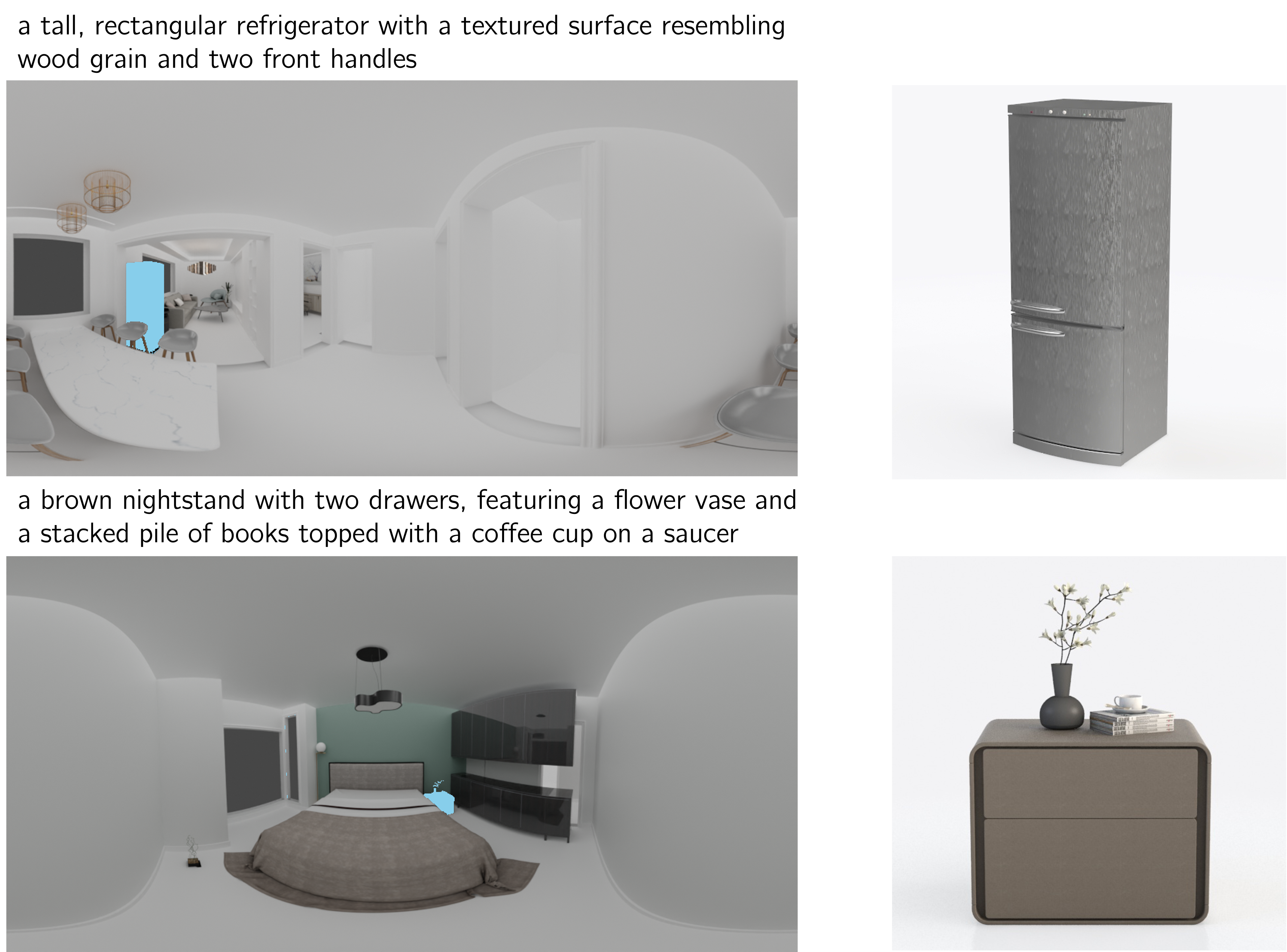}
    \caption{Samples from our dataset for the ROOMELSA challenge, part of SHREC 2025. Each example includes the rendered scene with the object masked in \textcolor{cyan}{cyan}, the corresponding text query provided to the retrieval model, and the ground-truth 3D mesh (e.g., refrigerator, nightstand) rendered from a canonical viewpoint.}
    \label{fig:dataset_overview}
\end{figure*}

\section{Related Work}\label{sec:related_work}
\subsection{Language-grounded Object Retrieval in 3D}
Language-grounded object retrieval in 3D has evolved from single-object detection to more complex set-level grounding. Specifically, ScanRefer~\cite{chen2020scanrefer} initiated the task by aligning 51,583 free-form referring expressions with 11,046 annotated objects across 800 \emph{ScanNet} reconstructions, requiring models to both \emph{detect} and \emph{select} an axis-aligned bounding box. In addition, ReferIt3D~\cite{achlioptas2020referit3d} addressed key limitations by providing ground-truth instance-level annotations and reformulating the task as a classification problem over candidate proposals, while also expanding coverage to 97,958 utterances across 707 scenes. Its division into templated (Sr3D) and natural (Nr3D) subsets highlights linguistic ambiguity rather than visual complexity as the primary source of early model failures. Extending the challenge further, Multi3DRefer~\cite{zhang2023multi3drefer} introduced 61,926 verified descriptions referencing zero, one, or multiple objects (11,609 instances) within the same 800 scenes. 
Collectively, these datasets establish a more rigorous benchmark suite for multimodal 3D scene understanding, advancing from coarse single-box retrieval to fine-grained, variable-cardinality grounding.

\subsection{Referring Segmentation}
Referring segmentation originated in the 2D, with datasets such as RefCOCO~\cite{kazemzadeh2014referitgame}, RefCOCO+~\cite{kazemzadeh2014referitgame}, and RefCOCOg~\cite{mao2016generation} pairing natural language expressions with object masks from MS-COCO~\cite{lin2014microsoft}. Early baselines, such as MAttNet\cite{yu2018mattnet}, introduced modular attention mechanisms that decompose each expression into components (e.g., \emph{subject}, \emph{location}, and \emph{relation}). More recently, transformer-based methods such as LAVT~\cite{yang2022lavt} and ReferFormer~\cite{wu2022language} have reformulated referring segmentation as a set prediction task, similar to Mask2Former~\cite{cheng2022masked}, thereby enabling more context-aware segmentation.

Building on these advances, language-driven part segmentation in 3D CAD environments has begun to be explored. PartNet~\cite{mo2019partnet} provides a large-scale benchmark with 573,585 part-level polygon annotations over 26,671 CAD models spanning 24 distinct object categories. In addition, Point2CAD~\cite{liu2024point2cad} proposes a hybrid analytic-neural reconstruction scheme that bridges segmented point clouds with structured CAD models and can be integrated with various segmentation backbones. However, such CAD-based settings are inherently free from real-world complexities, such as clutter, occlusion, and sensor noise, which limits their applicability to 3D perception tasks.

To overcome these limitations, recent work has shifted toward mask-level grounding in RGB-D scans, which combine the semantic richness of images with the geometric fidelity of real-world 3D data. This setting inherits the ambiguities of 2D vision while introducing additional challenges due to sparse and incomplete geometry. Mask3D~\cite{schult2023mask3d} adapts DETR-style mask queries to point clouds, achieving notable improvements in instance mAP on ScanNet200, although it relies on post-hoc language grounding via cosine similarity. In contrast, RefMask3D~\cite{he2024refmask3d} incorporates a dedicated language-processing branch alongside object-cluster tokens for tighter integration. Despite these improvements, most models remain limited to \emph{seen} scene categories during both training and evaluation. Open-vocabulary approaches such as SOLE~\cite{lee2024segment} seek to address this generalization gap by transferring segmentation priors from SAM into the 3D domain. However, they currently support only direct ``class-name $\leftrightarrow$ mask'' queries and do not model spatial relationships or compositional language, leaving a significant gap in expressive and flexible language grounding.

\subsection{Multimodal Foundation Models for 3D Understanding}
Contrastive Language–Image Pre-training (CLIP)~\cite{radford2021learning} demonstrated that training on 400 million noisy (image, text) pairs enables strong zero-shot transfer across a wide range of downstream tasks. Motivated by this success, recent efforts have extended the contrastive learning paradigm to 3D data. ULIP~\cite{xue2023ulip} aligns point cloud representations with the frozen CLIP embedding space, achieving competitive performance while significantly reducing the need for labeled data. However, ULIP and similar approaches typically embed entire visual \emph{instances} rather than fine-grained \emph{tokens}, leading to a phenomenon known as \emph{instance dispersion}: semantically coherent parts of the same object (e.g., the two legs of a chair) may be mapped farther apart in the embedding space than two distinct but visually similar objects, such as separate chairs. Significantly, this lack of part-level coherence undermines the consistency required for accurate instance-level mask prediction.

To address these limitations, ULIP-2~\cite{xue2024ulip}  introduces a tri-modal pre-training approach that combines geometry, vision, and language by generating comprehensive textual descriptions for 3D shapes using large multimodal models. By relying solely on 3D data as input, it eliminates the need for manual annotations, enabling efficient and scalable training on large datasets. In addition, Objaverse-XL\cite{deitke2023objaverse} have increased the volume of geometry used for pre-training, yet they still struggle with spatially grounded language queries. Moreover, to improve 3D language alignment, 3UR-LLM~\cite{xiong20253ur} introduces a pipeline that leverages open-source 2D vision-language models (VLMs) and large language models (LLMs) to automatically generate high-quality 3D–text pairs, culminating in the creation of the 3DS-160K dataset to enhance pre-training effectiveness.

\section{Problem Analysis and The Proposed Contributions}\label{sec:discuss}

\subsection{Limitations of Previous SHREC Challenges}
Over the past two decades, SHREC has significantly advanced the field of 3D retrieval through a series of focused subtasks, including image-to-shape matching~\cite{Hameed-SHREC2018,Hameed-SHREC2019,Li-SHREC2019,Li-SHREC2020,Feng-SHREC2022}, sketch-based querying~\cite{Li-SHREC2014,Juefei-SHREC2019,Qin-SHREC2022}, and point cloud recognition~\cite{le2023sketchanimar, le2023textanimar}. However, each track evaluates objects in isolation and relies on a single clean query modality. As a result, models are not exposed to real-world challenges such as cluttered environments, partial observability, or the need to integrate spatial context with free-form natural language. In addition, ground-truth annotations are often defined at the category level, and gallery sizes are relatively small, typically ranging from a few hundred to a few thousand shapes. Under these conditions, systems can achieve high performance by learning class-level templates without reasoning about instance-specific attributes. This limitation reduces the ecological validity of the evaluations and limits their applicability to more complex retrieval scenarios.

\subsection{Advantages of ROOMELSA Challenge}
In SHREC 2025, ROOMELSA is the first benchmark to formulate the task of \emph{scene-to-shape} retrieval. Given a whole, cluttered panoramic scene and a binary mask, the objective is to retrieve the exact CAD mesh that corresponds to the described object and is suitable for simulation or augmented reality overlay. This task unifies three traditionally distinct challenges: spatial grounding, linguistic disambiguation, and gallery-level retrieval. It evaluates whether a method can directly translate perception into an asset that downstream applications can utilize.

In addition, the dataset explicitly separates the localization step from semantic matching by providing both a spatial mask and an attribute-rich natural language description. This design compels models to resolve fine-grained object attributes such as color variations, part configurations, and material finishes, which become essential once the object’s location is known. ROOMELSA exposes the tie-breaking regime often absent in earlier tracks and underscores the importance of cross-modal alignment at the level of individual object properties.

Moreover, each query reflects a realistic scenario in augmented reality or robotics, where a user or agent observes a partially occluded object, marks its location, and requests a digital counterpart. Therefore, successful retrieval supports immediate downstream applications, including mesh substitution for physics simulation, virtual staging, or robotic manipulation. Notably, ROOMELSA encourages the development of retrieval systems that generalize beyond benchmark-specific heuristics and contribute to robust, deployable 3D understanding.

\section{Dataset and Evaluation}\label{sec:dataset}
In modern graphics and robotics pipelines, perception does not conclude with object detection alone. It concludes when a reusable asset is identified and ready for downstream tasks such as simulation, fabrication, or augmented reality rendering. This work addresses the whole pipeline in a single step by posing the following question: \emph{Can a model interpret a designer-style sentence, locate the described object within a cluttered 3D scene, and immediately return the exact \underline{3D model} that represents it?} We answer this question with ROOMELSA, a benchmark that combines dense scene-level masks with large-scale gallery-based mesh retrieval. Unlike earlier datasets focusing on bounding-box localization or part segmentation, ROOMELSA requires systems to ground language in scene context and resolve object identity at the instance level. In our ROOMELSA challenge, we introduce a novel \textit{mask-conditioned, language-driven} formulation for 3D object retrieval, with performance evaluated using Recall and the rank-sensitive Mean Reciprocal Rank. A qualitative dataset overview is presented in Fig.~\ref{fig:dataset_overview}.

\subsection{Scene Rendering}\label{sec:scene_rendering}
\begin{figure}[t]
    \centering
    \begin{subfigure}[b]{0.48\textwidth}
        \includegraphics[width=\linewidth]{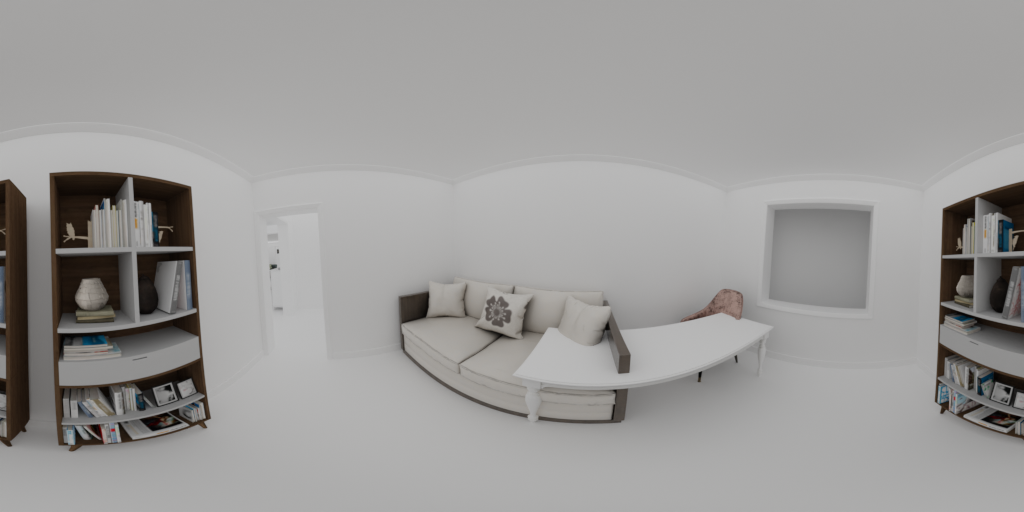}
        \caption{A rendered living room scene with diverse furniture placement and indirect lighting.}
    \end{subfigure}
    \hfill
    \begin{subfigure}[b]{0.48\textwidth}
        \includegraphics[width=\linewidth]{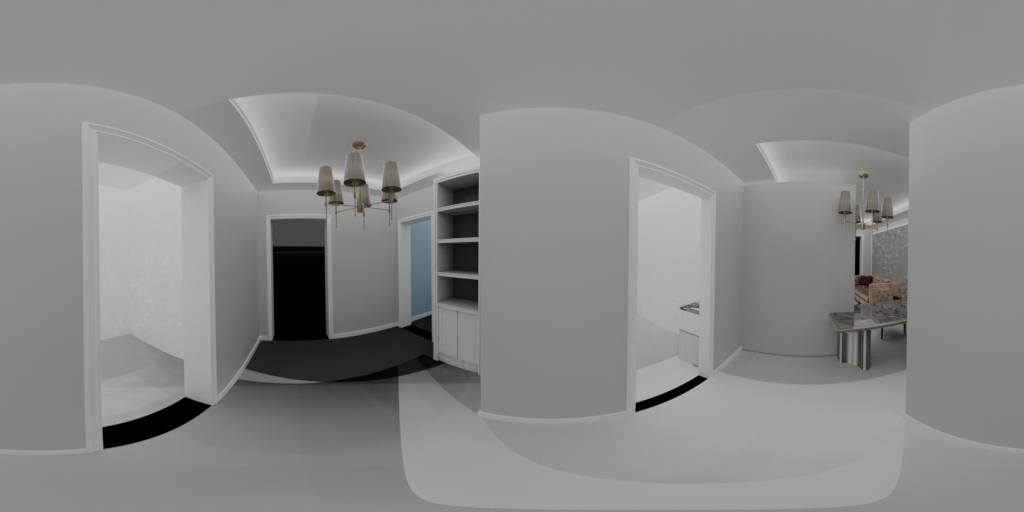}
        \caption{A rendered dining area with spatial compartments and complex object arrangements.}
    \end{subfigure}
    \caption{Examples of ROOMELSA-rendered scenes using \textsc{BlenderProc}. The photorealistic quality supports visual grounding and retrieval in our challenge.}
    \label{fig:rendered_scenes}
\end{figure}

In the ROOMELSA dataset, all panoramas are generated using \textsc{BlenderProc}, based on the original 3D-FRONT~\cite{fu20213d} layouts and 3D-Future~\cite{fu20213d} furniture meshes. For each selected apartment, we load the architectural shell, furniture models, and texture atlases using the official mapping file that links 3D-Future category IDs to 3D-FRONT semantic labels. The Cycles rendering engine is then configured to produce high-fidelity outputs by adjusting the global illumination parameters. In particular, the number of permitted light bounces is set to 200 across diffuse, glossy, transmission, and transparent interactions. This configuration helps reduce energy loss in visually complex areas such as drawers, lampshades, and other occluded cavities. Fig.~\ref{fig:rendered_scenes} presents rendered scenes, illustrating the photorealistic quality and spatial diversity achieved through this setup.

A single camera is placed in each recognized room. Camera poses are generated by sampling an unoccupied ``blank'' location 1.5 meters above the floor. A random yaw is drawn uniformly from the interval $[0, 2\pi)$, while the pitch is fixed at $90^\circ$ to ensure the equator of the panorama corresponds to human eye level. In rooms where no valid blank location is found, the sampler gracefully defaults to the next best available position.

Each panorama refers to a panoramic image of a room and is rendered at a resolution of $1024 \times 512$ pixels (internally specified with \texttt{width} $=$ 512 $\times$ 2 and \texttt{height} = 512). These images are path-traced under a consistent industrial HDR environment, identical to the one used during material transfer. In our implementation, a query typically consists of such a panorama, accompanied by a mask and a textual description.

In addition to RGB outputs, per-vertex instance and category ID maps are generated by enabling segmentation passes during the rendering process. Significantly, these panoramas in our dataset serve two purposes: (i) they provide photometric supervision for vision-based models, and (ii) they facilitate qualitative diagnostics when retrieval systems return incorrect CAD meshes, enabling analysis of whether errors stem from appearance, geometric mismatch, or contextual misunderstanding.

\subsection{Language Annotation Pipeline}\label{sec:language_annotation}

Our annotation process on the ROOMELSA dataset involves transforming a masked region into a designer-style sentence through a two-stage pipeline that combines automated generation with rigorous human quality control and verification.

\noindent\textbf{Stage 1 – Automated Annotation.} We leverage a vision–language model (e.g., LLaMA 3.2~\cite{touvron2023llama}) to process the RGB crop defined by the mask and generate a \emph{single, attribute-centric sentence}. The model is guided to begin with the object class and to describe the most visually salient attributes, including color, geometry, material, and distinctive parts, in compact prose. The model produces approximately 4,600 drafts per hour, providing complete coverage of all 44,000 masked objects with sufficient redundancy for downstream filtering. The model produces approximately 4,600 drafts per hour, ensuring coverage of the more than 50,000 masked objects referenced in the Introduction, with sufficient redundancy for downstream filtering. Sentence statistics confirm adherence to the desired style: the mean sentence length is $13.7 \pm 4.1$ tokens, and each sentence explicitly references 2–3 attributes, offering retrieval algorithms a rich yet concise semantic signal.

\noindent\textbf{Stage 2 – Human Refinement and Verification.}
A team of trained annotators reviews each machine-generated sentence, mask, and provisional CAD match. Annotators first refine the mask using a brush tool to eliminate boundary leakage. They then revise the sentence for factual accuracy, correcting, for example, misidentified surface finishes or structural details such as handle shapes. If necessary, they replace the provisional CAD ID with a more accurate match selected from a gallery of 55,000 models. A triple-agreement protocol is then applied: a (mask, sentence, model) triple is accepted only if three independent raters concur on both geometric fidelity and textual correctness. The finalized set achieves an inter-annotator IoU of $0.87 \pm 0.04$ and retains 97\% of the attribute tokens generated by the model, indicating that the automated drafts capture essential semantics while human edits focus on fine-grained accuracy.

This two-tier annotation pipeline yields 44,445 verified (mask, sentence, 3D model) triples on the ROOMELSA public and private test sets, balancing large-scale generation with the descriptive rigor required for reliable 3D model retrieval.

\subsection{Data Split}\label{sec:data_split}

The dataset is divided into two \emph{evaluation-only} partitions designed to support experimentation and leaderboard comparison.

\noindent\textbf{Public Test.} 
The primary partition includes 5,197 rooms drawn from 1,622 rendered apartments, each room paired with one or more fully verified (mask, sentence, 3D model) query triples, totaling 44,445 examples. This public test set provides ground-truth annotations for training, ablation, and local evaluation.

\noindent\textbf{Private Test.} For leaderboard evaluation, we manually sample \emph{50} diverse scenes, ensuring a balanced distribution across functional room types (e.g., bedrooms, kitchens, living rooms). From each selected scene, exactly \emph{one} masked object is chosen to form the private test split. Participants submit a ranked list of ten retrieval results per query, and evaluation is conducted server-side using Recall and MRR. We refer to this 50-query evaluation subset as the private test set throughout this paper.

Because the private test split contains exactly one query for each of its 50 distinct scenes, a single error lowers every metric (R\@1, R\@5, R\@10, and MRR) by precisely $1/50 = 0.02$. Therefore, this fixed decrement makes score changes straightforward to interpret, so even small gaps on the leaderboard (see Sec.~\ref{sec:results}) correspond to concrete gains or losses on individual queries.

\subsection{Evaluation Metrics}\label{sec:metrics}
We evaluate performance using four key metrics. The first three are Recall$@k$ metrics, which measure the proportion of queries for which the correct mesh appears within the top $k$ ranked predictions, where $k \in {1,5,10}$. These metrics provide a broad view of retrieval coverage, with higher values indicating that the correct model is more likely to appear among the top candidates. The fourth metric is Mean Reciprocal Rank at 10, which quantifies the rank position of the correct mesh within the top 10 predictions. Its formal definition is provided in Eqn.~\eqref{eqn:mrr}.
\begin{equation}
    \text{MRR} = \frac{1}{|Q|}\sum_{i=1}^{|Q|} \frac{1}{\operatorname{rank}_i}
    \label{eqn:mrr}
\end{equation}
where $|Q|$ is the total number of queries, and $\operatorname{rank}_i$ is the position of the correct model in the ranked list for the $i$-th query. If the correct model does not appear among the top predictions, $\operatorname{rank}_i$ is set to $\infty$, which contributes a final score of zero.

While Recall focuses on whether the correct mesh is retrieved, MRR emphasizes the importance of its position in the ranked list. For example, moving a correct model from rank 2 to rank 1 doubles its contribution to the final score. By reporting both families of metrics, ROOMELSA allows for the distinction between retrieval coverage (Recall) and ranking precision, providing a clearer understanding of the model's performance.

\section{Participants}
\label{sec:participants}
\begin{figure*}[t]
    \centering
    \begin{minipage}{0.49\textwidth}
        \centering
        \includegraphics[width=\linewidth]{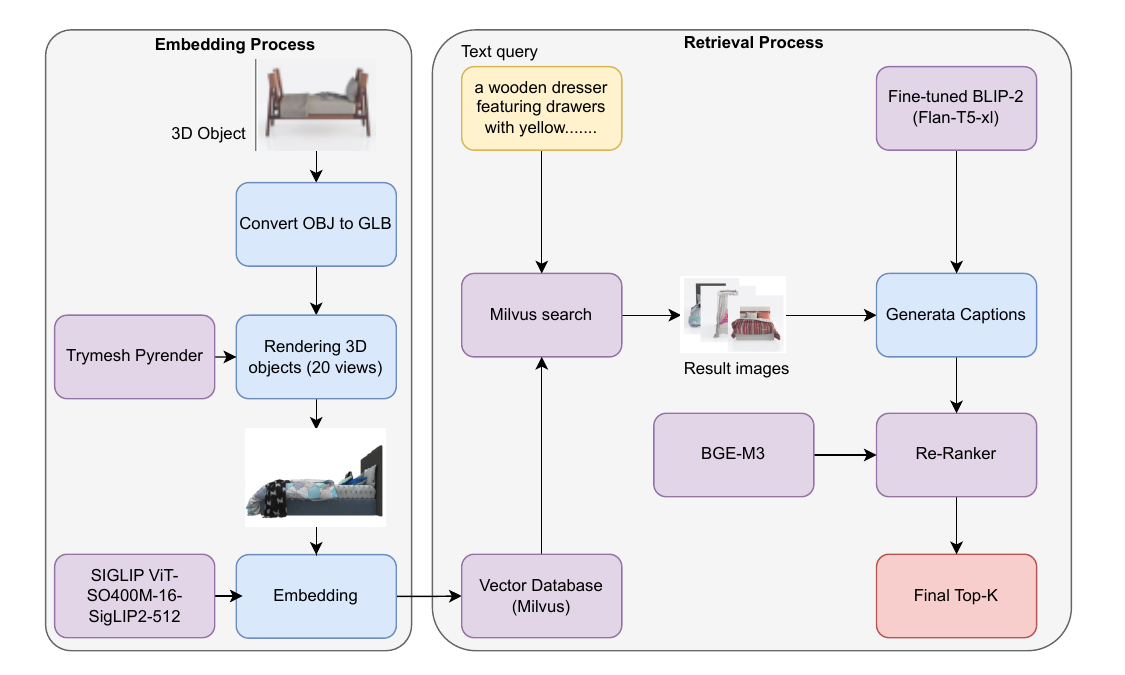}
        \caption*{(a) The pipeline of team Stubborn\_Strawberries (\textcolor{blue}{\textit{the winning team of the ROOMELSA challenge}}). In particular, SIGLIP embeddings provide rapid coarse search; BLIP-2 captions and BGE-M3 similarities supply semantic reranking.}
    \end{minipage}
    \hfill
    \begin{minipage}{0.49\textwidth}
        \centering
        \includegraphics[width=\linewidth]{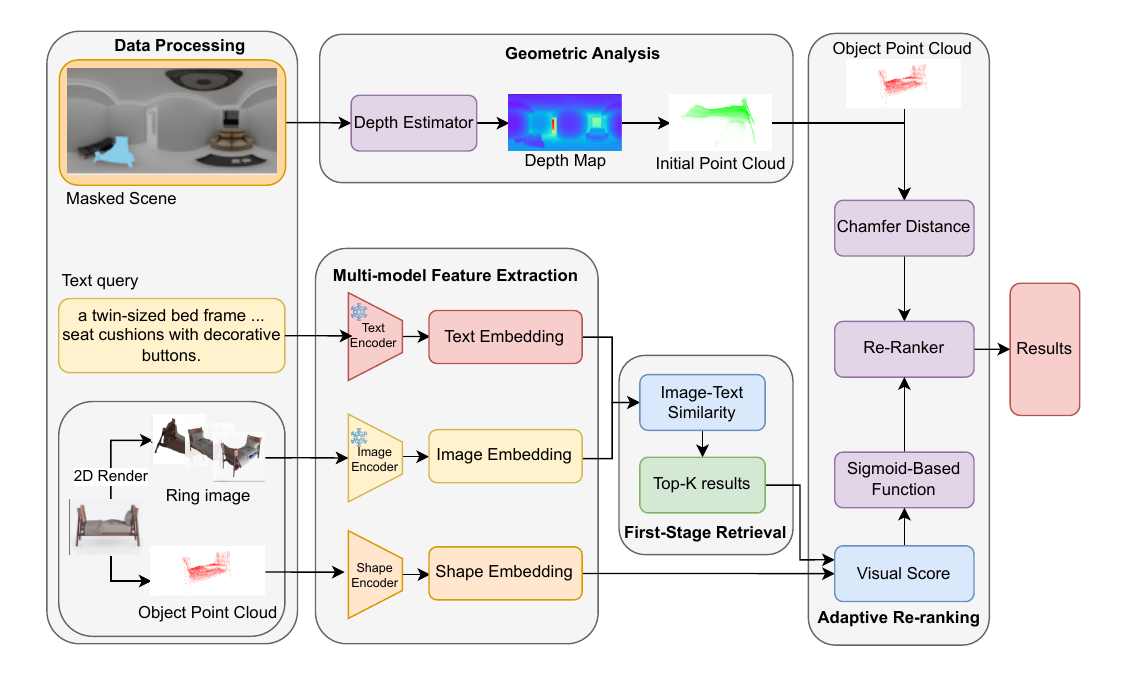}
        \caption*{(b) The pipeline of team Ai-Yahh. Specifically, the proposed COMPASS method incorporates multi-view CLIP embeddings, PointBERT geometry, and depth-based query reconstruction, along with adaptive cross-modal re-ranking.}
    \end{minipage}
    \caption{Overview of the proposed method introduced by two teams: Stubborn\_Strawberries, and Ai-Yahh.}
    \label{fig:methods_1}
\end{figure*}
The inaugural ROOMELSA evaluation attracted \emph{18 participating teams}, including university laboratories and research groups from across Asia and Africa. Following the execution of their open-source solutions on the private 50-query test split (Sec.~\ref{sec:data_split}), the organizers compiled an official leaderboard and selected the \emph{top five} performers for further analysis in Sec.~\ref{sec:results}.
\begin{enumerate}
    \item \textbf{Stubborn\_Strawberries} comprises \textit{Long Le Bao, Thai Hoang Minh, Minh Nguyen Anh, Thang Nguyen Tien, Phat Nguyen Thuan, Huy Nguyen Phong, Bao Huynh Thai, Vinh-Tiep Nguyen, and Duc-Vu Nguyen}, collaborating in the Multimedia Laboratory. By leveraging complementary expertise in 3D perception and multimodal retrieval, the team achieved the highest overall score in the challenge, establishing the ROOMELSA benchmark (see Sec.~\ref{team:Stubborn_Strawberries}).
    
    \item \textbf{Ai-Yahh} consists of three members: \textit{Phu-Hoa Pham, Minh-Huy Le-Hoang, and Nguyen-Khang Le}. The team proposed an engineered solution with new ideas and achieved second place on the leaderboard (see Sec .~\ref {team:Ai-Yahh}).
    
    \item \textbf{MealsRetrieval}, composed of \textit{Minh-Chinh Nguyen, Minh-Quan Ho, Ngoc-Long Tran, Hien-Long Le-Hoang, Man-Khoi Tran, and Anh-Duong Tran}, drew on a multidisciplinary background spanning computer graphics, natural language processing, and human–computer interaction. This expertise contributed to their top-three (see Sec.~\ref{team:MealsRetrieval}).
    
    \item \textbf{BUCCI\_GANG} is a collaborative team comprising members from multiple national universities, including \textit{Kim Nguyen, Quan Nguyen Hung, Dat Phan Thanh, Hoang Tran Van, Tien Huynh Viet, and Nhan Nguyen Viet Thien}. Their submissions provided an efficient method across the diverse scenes and secured fourth place (see Sec.~\ref{team:BUCCI_GANG}).
    
    \item \textbf{NoResources}, consisting of \textit{Dinh-Khoi Vo, Van-Loc Nguyen, and Trung-Nghia Le} from the Software Engineering Laboratory, secured the final spot in the ROOMELSA top five. Their resource-efficient approach highlights that carefully designed models can effectively compete with more computationally intensive solutions (see Sec.~\ref{team:NoResources}).
\end{enumerate}

The five teams with the highest aggregate scores across Recall and MRR established the performance baseline for ROOMELSA. A detailed comparison of their proposed methods is provided in Sec.~\ref{sec:methods}, with corresponding results reported in Sec.~\ref{sec:results}. These methods were selected based on their performance on the leaderboard.  Notably, \textit{none of the ROOMELSA organizers participated in the challenge or submitted results}.

\section{Methods}
\label{sec:methods}
\begin{figure*}[t]
    \centering
    \begin{minipage}{0.49\textwidth}
        \centering
        \includegraphics[width=\linewidth]{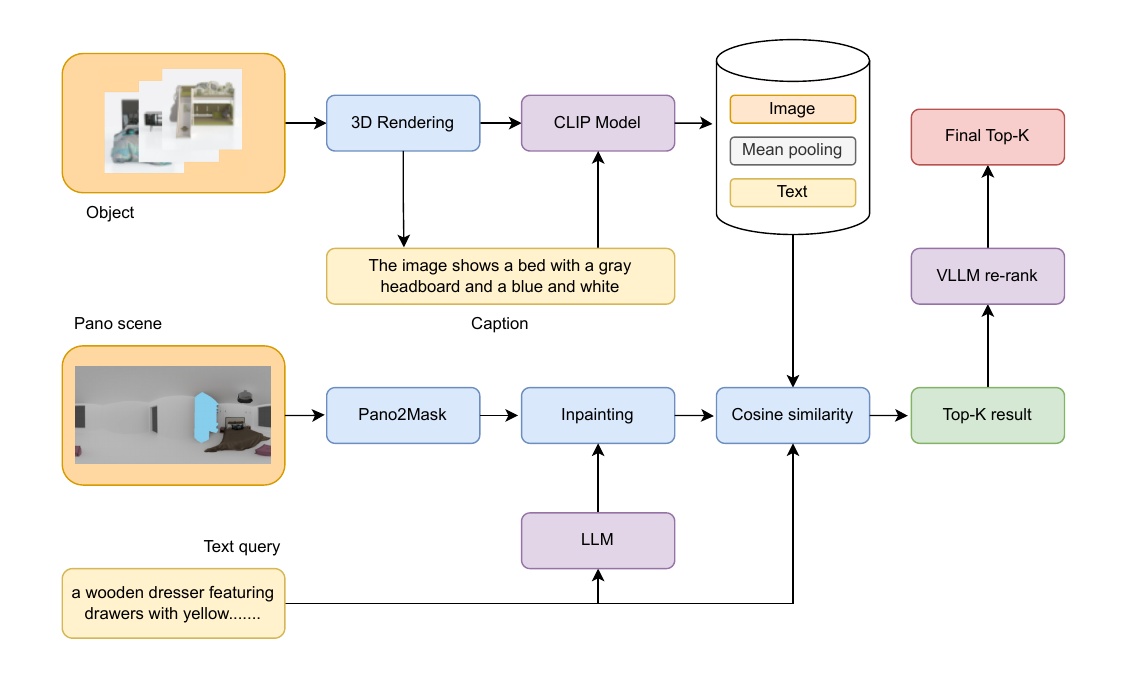}
        \caption*{(a) The proposed method from team MealsRetrieval. Meshes are rendered and captioned offline. During inference, the query panorama is masked, inpainted, and embedded. In addition, a dual-stream CLIP search yields ten candidates.}
    \end{minipage}
    \hfill
    \begin{minipage}{0.49\textwidth}
        \centering
        \includegraphics[width=\linewidth]{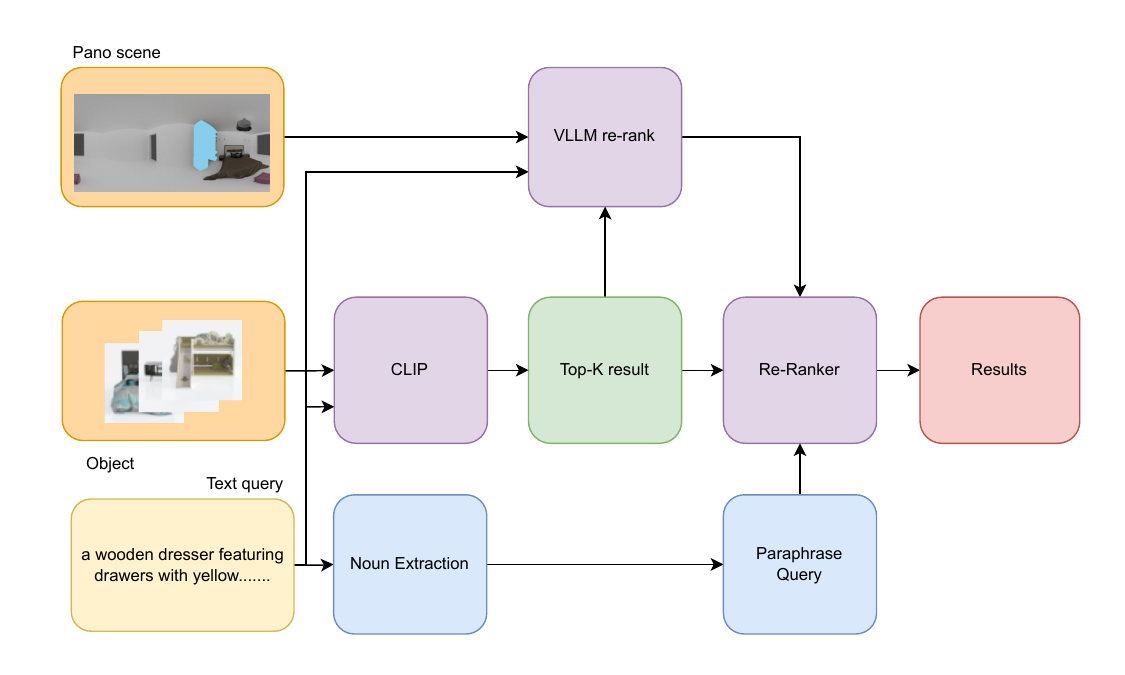}
        \caption*{(b) The proposed method from team BUCCI\_GANG. Specifically, a CLIP cosine search delivers a top-10 shortlist, after which InternVL-2.5-2B scores each candidate in the context of the masked panorama and the refined query.}
    \end{minipage}
    \vspace{-1cm}
    \begin{minipage}{0.7\textwidth}
        \centering
        \includegraphics[width=\linewidth]{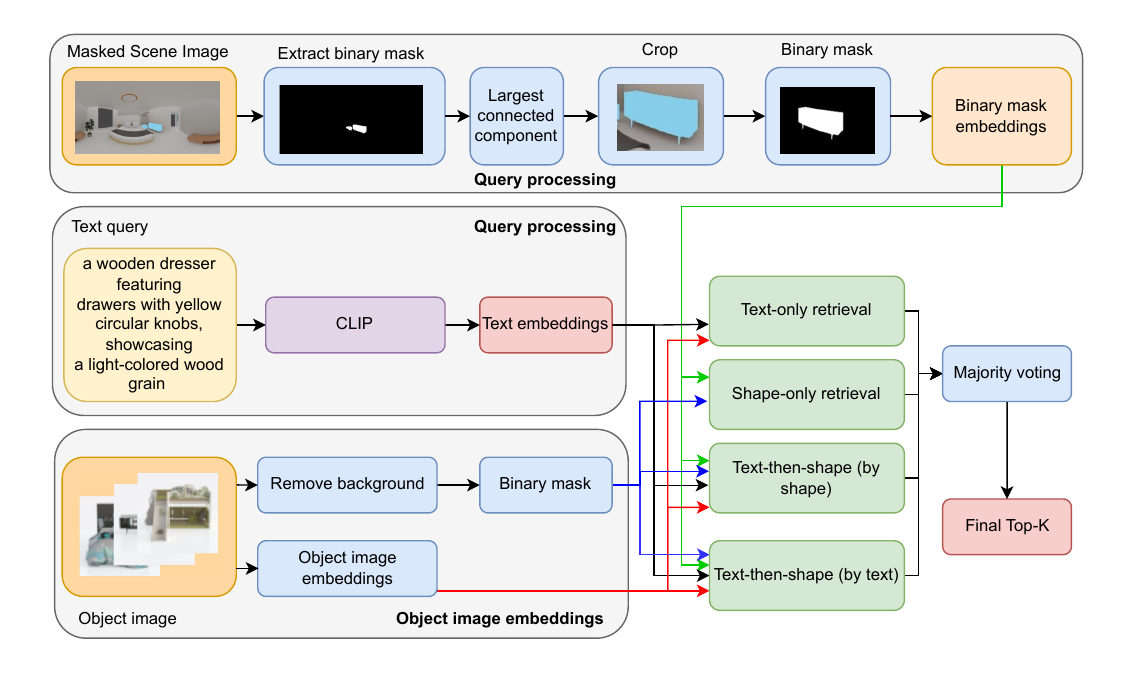}
        \caption*{(c) The proposed method from team NoResources. Specifically, the proposed method leverages binary mask extraction, dual CLIP embeddings for RGB and silhouette, five retrieval strategies, and a simple weighted vote fusion.}
    \end{minipage}
    \vspace{2\baselineskip}
    \caption{Overview of the proposed method developed collaboratively by three teams: MealsRetrieval, BUCCI GANG, and NoResources.}
    \label{fig:methods_2}
\end{figure*}
\noindent\textbf{Overview of Submitted Solutions.} The submitted solutions span a broad spectrum, ranging from lightweight, zero-finetuning baselines to complex multimodal cascades that integrate appearance, geometry, depth, and large language model priors. Specifically, approximately half of the teams rely solely on frozen CLIP-style encoders combined with carefully designed post-processing steps. The remaining teams incorporate additional cues such as BLIP-generated captions, PointBERT-based point clouds, depth panoramas, or scene-level reasoning using vision and language models to address the challenge of fine-grained ranking. Despite this methodological diversity among teams, all high-performing pipelines adopt the same overarching structure. Each begins with a fast vector search that ensures perfect top-10 recall, followed by a semantics-aware re-ranker that determines the final ranking. This convergence across varied approaches highlights the effectiveness of the two-stage retrieval paradigm and provides a clear blueprint for future research in language-driven 3D model retrieval.

\subsection{Team Stubborn\_Strawberries}
\label{team:Stubborn_Strawberries}

Fig.~\ref{fig:methods_1}a shows the winning pipeline proposed by \emph{Stubborn\_Strawberries}, which follows a two-stage process. The first stage employs a high-speed vector search based on multi-view visual embeddings, while the second stage applies a language-aware re-ranking module to resolve ambiguities.

\noindent\textbf{Stage 1 – Multi-view Embedding and Coarse Retrieval.} Each gallery mesh is rendered uniformly from 20 azimuthal angles (at 18$^{\circ}$ intervals, camera radius 3m) at a resolution of $1024 \times 1024$ pixels, under a white light dome. Every view is encoded into a 1,152-dimensional embedding using SIGLIP~\cite{zhai2023sigmoid} (ViT-SO400M-16–SigLIP2-512, pre-trained on WebLI). The resulting embeddings are stored in a \texttt{FLAT} index within Milvus, using cosine distance for similarity search. At query time, the masked crop from the input scene is processed using the same SIGLIP encoder, and a $k = 10$ nearest-neighbor search retrieves an initial shortlist of candidate meshes. The end-to-end latency is under 50ms, rendering it negligible within the framework.

\noindent\textbf{Stage 2 – Language-Guided Reranking.}
The ten candidate meshes are refined in two steps. First, a BLIP-2 captioning model (with a Flan-T5-XL~\cite{chung2024scaling} decoder, fine-tuned for 20 epochs on 3D renderings) generates a one-sentence textual description for each mesh. During this process, the vision encoder is frozen while the language parameters are updated. Second, the original query and the generated captions are embedded using BGE-M3~\cite{xiao2024c} (bge-large-en-v1.5), and cosine similarity is used to rerank the candidates. This promotes meshes whose self-described attributes most closely align with the user's input sentence. In addition, this lightweight reranker effectively resolves fine-grained distinctions (e.g., selecting a ``grey two-drawer nightstand'') over a visually similar three-drawer dresser when the query explicitly references the number of handles.

\emph{Stubborn\_Strawberries} execute this proposed method on a single A100-80GB GPU. In particular, SIGLIP and BGE-M3 are used off-the-shelf, while BLIP-2~\cite{li2023blip} is the only component subjected to fine-tuning. In addition, an ablation study provided by the team demonstrates that reranking improves Recall\@1 by 14 percentage points over vector search alone, confirming the benefit of explicit language alignment in the retrieval process.

\subsection{Team Ai-Yahh}\label{team:Ai-Yahh}

Fig.~\ref{fig:methods_1}b outlines COMPASS, the second-ranked system submitted by team \emph{Ai-Yahh}.  
In particular, the proposed COMPASS method fuses appearance, geometry, and depth cues in a three-stage cascade:  
(i) multi-view feature extraction,  
(ii) 3-D representation learning, and  
(iii) geometry-aware re-ranking.

\noindent\textbf{Multi-view appearance features.}
Each gallery mesh is rendered from twelve viewpoints, three elevation angles (\(\pm30^{\circ}\) and \(0^{\circ}\)) crossed with four azimuths (45°, 135°, 225°, 315°). Specifically, CLIP ViT BigG-14 embeds every image, pre-trained on two billion image–text pairs. In addition, the twelve vectors are averaged to form a 768-D appearance descriptor, \(\mathbf{f}_{\text{CLIP}}\).

\noindent\textbf{Point-level geometry features.}
Geometric information is captured with a PointBERT~\cite{yu2022point} encoder borrowed from OpenShape~\cite{liu2023openshape}. The network is fine-tuned in two phases:  40 epochs on 11k public ROOMELSA meshes, followed by 10 epochs on the entire gallery, each time sampling 10k points per object. In addition, the resulting 1,024-D embedding, \(\mathbf{f}_{\text{PB}}\), complements the appearance cue especially for texture-poor shapes.

\noindent\textbf{Depth-guided query reconstruction.}
For every masked query, the authors predict a dense depth panorama with PanoFormer~\cite{shen2022panoformer}. Specifically, pixels inside the mask are back-projected to 3-D using spherical coordinates,
\(\mathbf{p}(u,v)=r(u,v)\,[\sin\theta\cos\varphi,\;\sin\theta\sin\varphi,\;\cos\theta]\),  
yielding an \emph{initial} object point cloud \(\mathcal{P}_{\text{init}}\).  
Although partial, \(\mathcal{P}_{\text{init}}\) suffices to compute Chamfer distances \(d_{\text{CD}}\) to candidate meshes.

\noindent\textbf{Adaptive cross-modal re-ranking.} Finally, the final similarity score mixes three terms, which is formally defined in Eqn.~\eqref{eqn:compass}.
\begin{equation}
\begin{aligned}
S_{\text{final}}
=\, & w_{\text{adapt}} \bigl(S_{\text{CLIP}} + S_{\text{PB}}\bigr) \\
& + (1 - w_{\text{adapt}})\bigl(S_{\text{CLIP}} + S_{\text{PB}} + w_{\text{CD}}\,S_{\text{CD}}\bigr)
\end{aligned}
\label{eqn:compass}
\end{equation}
where \(S_{\text{CLIP}}\) and \(S_{\text{PB}}\) are cosine similarities in the two embedding spaces,  
\(S_{\text{CD}}\) converts Chamfer distance into a bounded similarity via a smoothed exponent,  
and \(w_{\text{adapt}}=\bigl[1+e^{-10(S_{\text{CLIP}}+S_{\text{PB}}-0.13)}\bigr]^{-1}\)  
automatically down-weights geometry whenever the visual match is already confident.  


\subsection{Team MealsRetrieval}\label{team:MealsRetrieval}

\emph{MealsRetrieval} converts the ROOMELSA task into a
multi-modal \emph{2-D image + text} retrieval problem
(overview in Fig.~\ref{fig:methods_2}a).
The system comprises four blocks:
multi-view rendering and captioning of the gallery meshes,
Panorama-to-mask extraction,
LLM-guided inpainting of the query object,
and a dual-stream retrieval module followed by
VLLM-based re-ranking.

\noindent\textbf{Gallery Preparation.} Each 3-D mesh is surrounded by two elevation rings
(\(\pm30^{\circ}\)) and imaged every \(30^{\circ}\) in azimuth,
plus nadir and zenith views, for a total of 28 renderings
at \(1024\times1024\) px.
Florence-2~\cite{xiao2024florence}, large generates a single-sentence caption
from an oblique view (\(30^{\circ}{-}45^{\circ}\) elevation),
which empirically yields the highest attribute recall.
In particular, CLIP embeds every image (ViT-L/14) and the caption and vectors are stored in a Faiss index together with a
mean-pooled ``object fingerprint'' that stabilizes retrieval
from unseen viewpoints.

\noindent\textbf{Panorama-to-mask.} Given a scene panorama and the text query, Pano2Mask enumerates planar crops by spherical projection, runs the YOLOv11~\cite{khanam2024yolov11} model to detect candidate boxes,
and ranks them by ``centre-object proximity'' and
dominant-color conformity. The crop whose masked pixels best match the query color becomes the working view. Additionally, non-target pixels are whitened to produce a clean, binary mask.

\noindent\textbf{Query Inpainting.} A bilingual LLM (e.g., EXAONE-3.0-Instruct~\cite{an2024exaone}) rewrites the raw query into a concise, disambiguated prompt. In particular, Zero-Painter~\cite{ohanyan2024zero} fills the masked region, conditioned on the refined sentence, yielding a complete RGB image that CLIP can embed. Notably, this step proves essential for oddly-shaped or heavily occluded objects, where a partial crop would mislead the visual
encoder.

\noindent\textbf{Dual-stream Retrieval and Re-ranking.}
If the mask quality score lies in \([2200,\,100\,000]\)
(the “medium’’ band),
retrieval is driven by the \emph{text} embedding;
otherwise the \emph{image} embedding is used.
In addition, a cosine search then returns the top ten candidates,
which are passed through the InternVL2\_5-MPO~\cite{wang2024enhancing} model for semantic re-ranking. Moreover, the multimodal LLM compares each candidate image with the original query and promotes the two meshes that best match the linguistic intent.


\subsection{Team BUCCI\_GANG}\label{team:BUCCI_GANG}
Fig.~\ref{fig:methods_2}b introduces the fourth-ranked entry from \emph{BUCCI\_GANG}. In particular, the proposed method keeps all backbone weights frozen and relies on a two-step strategy: (1) a CLIP-based retrieval module that delivers a compact candidate list, and (2) a vision-language model (VLLM) then re-examines each candidate in the full panoramic context.

\noindent\textbf{CLIP Shortlist.}
Every gallery rendering is embedded offline with OpenCLIP ViT-SO400M-14-SigLIP-384~\cite{cherti2023reproducible}. At run time, the user sentence is first sanitized by a dependency parser that removes opinion adjectives and purpose clauses, leaving a terse noun phrase. In addition, several paraphrases of that phrase are generated to widen lexical coverage. The same CLIP text encoder encodes the original query and its paraphrases, and a cosine search retrieves a top-\(k\) (default \(k=10\)) shortlist.

\noindent\textbf{Context-aware Re-ranking with a Frozen VLLM.}
Each shortlisted mesh image is paired with the masked panorama and fed to InternVL-2.5-2B~\cite{chen2024expanding}, a frozen vision–language LLM that outputs a scalar plausibility score.  
The score reflects stylistic harmony, functional fit, and geometric consistency, signals that simple feature distances cannot capture.  
CLIP similarity and VLLM plausibility are min–max normalized and fused linearly to produce the final ranking. This single pass eliminates most ambiguities left after the appearance-driven retrieval.


\subsection{Team NoResources}\label{team:NoResources}

Fig.~\ref{fig:methods_2}c provides an overview of the
\emph{NoResources} pipeline, which is a purely CLIP-based solution and relies on clever query processing and voting rather than heavy
fine-tuning. 

\noindent\textbf{Query Preprocessing.} The masked panorama is converted to a binary mask by thresholding the sentinel RGB color (135, 206, 235). Specifically, connected-component analysis retains only the largest region. The bounding box is dilated by 10 pixels to avoid truncation, and the masked crop is rebinarized to correct discretization gaps. Finally, the resulting RGB crop and its binary silhouette feed two separate embedding streams.

\noindent\textbf{Dual Embeddings for Every Catalogue Mesh.}
For each gallery object, they store  (i) an RGB embedding that captures color and texture, and  (ii) a binary-silhouette embedding that isolates pure shape. Both embeddings come from the same frozen CLIP~\cite{radford2021learning} encoder (ViT-L/14).  Backgrounds are removed using \texttt{rembg} before silhouette extraction, ensuring that contextual pixels do not pollute the geometric information.

\noindent\textbf{Five Complementary Retrieval Strategies.} Five distinct strategies for combining textual and visual information. (1) \emph{Text only} ranks candidates based on cosine similarity between the query sentence and RGB image embeddings. (2) \emph{Shape only} uses cosine similarity between the silhouette of the masked crop and precomputed object silhouettes. (3) \emph{Text-then-shape (shape order)} first filters the top 15 candidates by text similarity and then re-ranks the top 10 based on shape similarity. (4) \emph{Text-then-shape (text order)} applies the same filtering but retains the text-based order in the final ranking. (5) \emph{Majority vote} aggregates results from the previous strategies using a weighted voting scheme, assigning $+2$ to each text-then-shape variant and $+1$ to the remaining methods to produce the final ranked list.


\section{Results and Discussions}\label{sec:results}
\subsection{Quantitative Results}\label{sec:quan_res}

\noindent\textbf{Leaderboard Analysis.}
Table~\ref{tab:leaderboard} shows that five top-ranked models achieve perfect scores of $R@5 = R@10 = 1.00$, meaning each retrieves the correct mesh within the top-10 results for every query in the private test. This outcome suggests that combining rich textual descriptions and modern language embedding models effectively solves the coarse localization problem.

With coarse retrieval saturated, final ranking performance depends entirely on fine-grained ordering. Since the private test split contains exactly 50 queries, demoting the correct mesh from rank 1 to rank 2 reduces each evaluation metric by precisely $1/50 = 0.02$. Specifically, the 0.03 spread in MRR between \emph{Stubborn\_Strawberries} (0.97) and \emph{NoResources} (0.93), therefore, reflects just three additional scenes where the former placed the correct object first rather than second. In effect, the competition becomes a tie-breaking task, driven by linguistic subtleties, material distinctions, and part-level geometric cues.

The winning entry, \emph{Stubborn\_Strawberries} (Sec.~\ref{team:Stubborn_Strawberries}), integrates multi-view SIGLIP embeddings, BLIP-2-generated captions, and BGE-M3 sentence similarity with an adaptive weighting scheme, giving the reranker sufficient flexibility to resolve nearly all near-miss cases. In contrast, the next two teams reached comparable performance through orthogonal strategies: \emph{Ai-Yahh’s COMPASS} (Sec.~\ref{team:Ai-Yahh}) incorporates depth-aware point-cloud comparisons, while \emph{MealsRetrieval} (Sec.~\ref{team:MealsRetrieval}) leverages inpainted 2D views evaluated by a multimodal large language model. Their parity suggests that appearance-plus-geometry and appearance-plus-language pipelines are equally effective for achieving high precision.

In addition, \emph{BUCCI\_GANG} (Sec.~\ref{team:BUCCI_GANG}) demonstrates that a frozen vision–language model can close much of the remaining gap by producing scene-level plausibility scores. Meanwhile, \emph{NoResources} (Sec.~\ref{team:NoResources}) shows how far a lightweight ensemble of RGB images and silhouette-based CLIP embeddings can go when paired with thoughtful query sanitization.

Importantly, the near-perfect results achieved by the top five teams demonstrate that current models are highly effective at retrieving the correct object within the top-ranked candidates. However, the small differences in MRR, particularly between the first- and fifth-ranked teams, highlight that the real challenge lies in fine-grained ranking rather than basic retrieval. For example, the difference in MRR (0.97 vs. 0.93) reflects only a slight variation in the models' ability to place the correct object at the very top of the list, suggesting that the top-performing systems have largely mastered coarse localization. This narrow performance gap indicates that while current methods excel at narrowing down candidate objects, achieving perfect retrieval still depends on making subtle distinctions, such as identifying fine-grained differences in appearance, material, or part configuration. The tight score range further reinforces that, for most methods, the difficulty lies not in retrieving the correct object but in ranking it first with consistent precision.

\subsection{Submission Performance Trends of Top Teams}\label{sec:trajectories}
\begin{figure}[t]
    \centering
    \includegraphics[width=\linewidth]{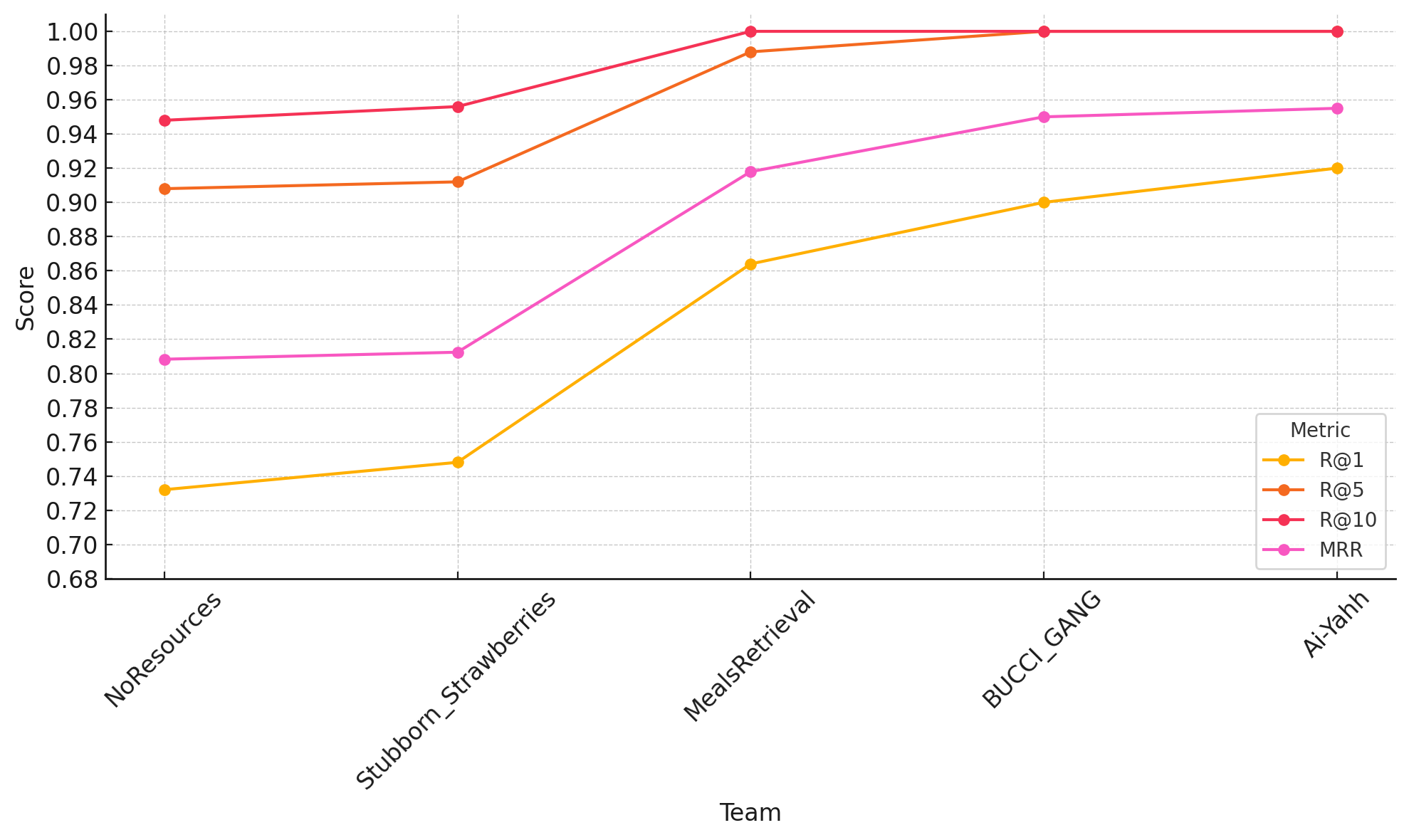}
    \caption{Performance trend of the top five teams over their five latest submissions.}
    \label{fig:submissions}
\end{figure}

To understand how teams iteratively converged on their final solutions, Fig.~\ref{fig:submissions} summarizes all submissions received during the three-week evaluation window. Each team was permitted up to five uploads; if fewer than five were submitted, the last available result was carried forward for comparison.

\emph{Ai-Yahh} and \emph{BUCCI\_GANG} achieved their final scores with their first submission and made no subsequent changes. This suggests that both teams conducted extensive offline validation prior to engaging with the leaderboard. In contrast, \emph{Stubborn\_Strawberries} began with an exploratory baseline (MRR $\approx$ 0.33), followed by two major performance gains: first, the introduction of multi-view SIGLIP raised MRR by 0.56; second, the addition of the BLIP-2 + BGE-M3 reranker contributed a further 0.09 improvement. \emph{MealsRetrieval} exhibited a similar two-phase trajectory, linked to enabling and disabling their inpainting-based LLM reranker. \emph{NoResources}, in contrast, showed consistent improvement with each submission, reflecting the cumulative effect of their modular ensemble design.

The flat submission curves of \emph{Ai-Yahh} and \emph{BUCCI\_GANG} indicate that their performance ceilings were constrained not by training instability, but by the representational limits of their frozen backbone models. This aligns with earlier observations that scene-level plausibility scoring, central to their pipelines, tends to resolve marginal cases rather than address major failures. Conversely, the sharp early gains by \emph{Stubborn\_Strawberries} highlight the critical role of multi-view rendering. A single viewpoint in their initial attempt left significant blind spots, while rendering each mesh from 20 views reduced retrieval ambiguity even before semantic reranking was applied.

These trajectories conclude that the current bottleneck lies in \emph{fine-grained ranking}. Teams that already encode sufficient geometric and appearance diversity tend to see diminishing returns from the leaderboard. In contrast, teams initially lacking such diversity benefit the most from multi-view augmentation and specialized re-rankers. Future iterations may consider increasing the difficulty by adding more visually similar object categories with subtle part-level differences or reducing the top-$k$ allowance to enhance the impact of ranking precision.

\begin{table}[t]
\centering
\caption{Official ROOMELSA leaderboard on the private test split.}
\label{tab:leaderboard}
\small
\begin{tabular}{lcccc}
\hline
\textbf{Team} & \textbf{R@1} & \textbf{R@5} & \textbf{R@10} & \textbf{MRR} \\ \hline
Stubborn\_Strawberries & \textbf{0.94} & \textbf{1.00} & \textbf{1.00} & \textbf{0.97} \\
Ai-Yahh                & 0.92 & 1.00 & 1.00 & 0.96 \\
MealsRetrieval         & 0.92 & 1.00 & 1.00 & 0.96 \\
BUCCI\_GANG            & 0.90 & 1.00 & 1.00 & 0.95 \\
NoResources            & 0.88 & 1.00 & 1.00 & 0.93 \\ \hline
\end{tabular}
\end{table}
\subsection{Qualitative Results}\label{sec:qual_res}

Among the five top-ranked pipelines, we visualise retrieval behavior for COMPASS (from \emph{Ai\_Yahh}) and \emph{NoResources} because these two systems bracket the design spectrum. In particular, COMPASS represents a heavy multimodal fusion of image, geometry, and depth. In contrast, the proposed method from NoResources is a compute-light ensemble built from frozen CLIP embeddings. Thus, their successes and failures reveal complementary insights into the ROOMELSA task.

\begin{figure*}[t]
    \centering
    \includegraphics[width=\linewidth]{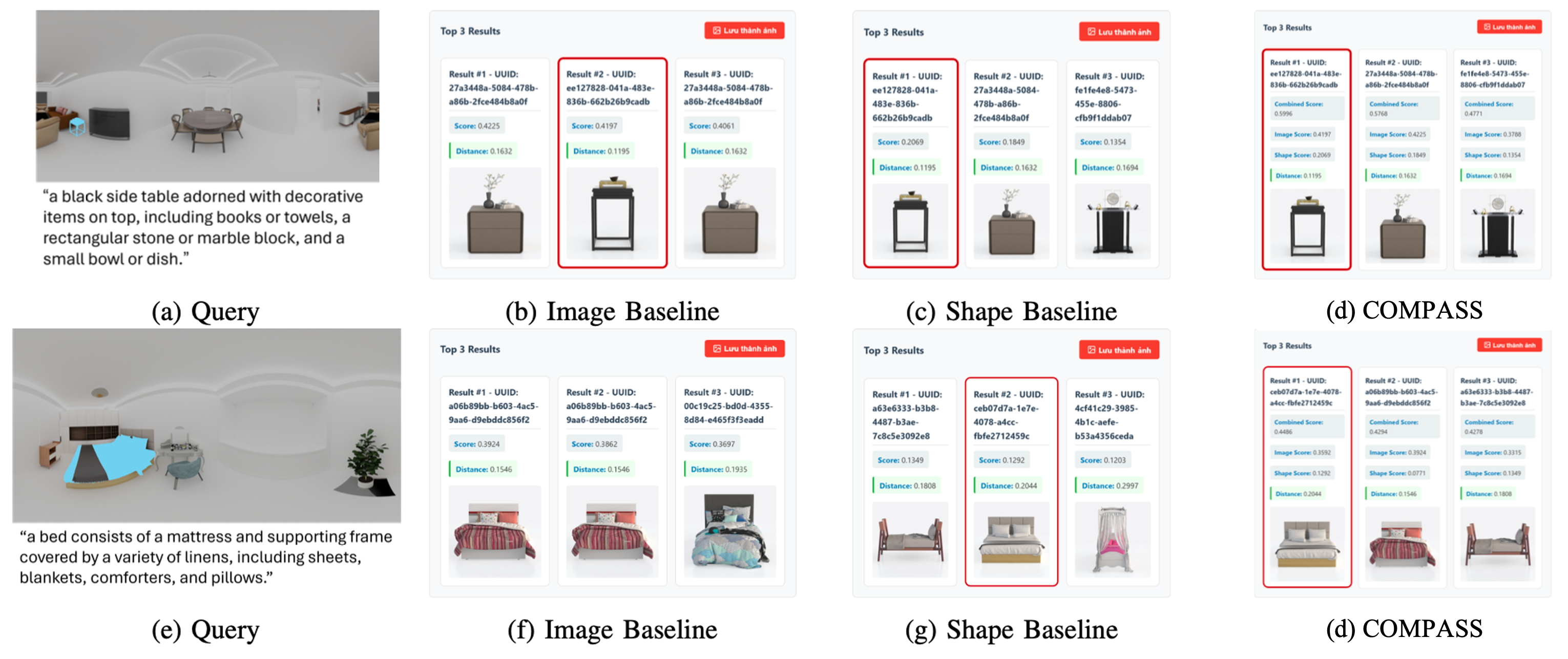}
    \caption{Qualitative comparison of COMPASS (from Ai\_Yahh) against two baselines. From left to right: the input scene with the masked query, the top retrieval result from the image-only baseline (CLIP on RGB), the top result from the shape-only baseline (geometry-based embedding), and the prediction from COMPASS.}
    \label{fig:vis_AiYahh_COMPASS}
    \vspace{\baselineskip}
\end{figure*}

Fig.~\ref{fig:vis_AiYahh_COMPASS} demonstrates how COMPASS outperforms two simplified baselines: an image-only variant that uses CLIP embeddings of RGB renderings, and a shape-only variant that relies solely on geometric similarity from point cloud embeddings. By integrating visual, geometric, and linguistic cues, COMPASS achieves more accurate rankings across a diverse range of queries. In both query examples, COMPASS correctly identifies the target object, highlighting the robustness of multimodal fusion in resolving visually similar distractors and ambiguous textual cues. In the upper panel, the query describes a black tripod side table with structural and decorative attributes. The image-only baseline ranks a similar but undecorated table first, while the shape-only baseline retrieves the correct tripod but misranks distractors. COMPASS combines image and PointBERT shape scores to resolve both issues and ranks the correct mesh first. In the lower panel, the query specifies both a slatted headboard and white linens. The image-only method overemphasizes fabric color, while the shape-only method prioritizes structure and neglects upholstery. COMPASS balances aspects, retrieving the only mesh that satisfies the full description, demonstrating its effectiveness in complex scenarios.

\begin{figure*}[!t]
    \centering
    \includegraphics[width=\linewidth]{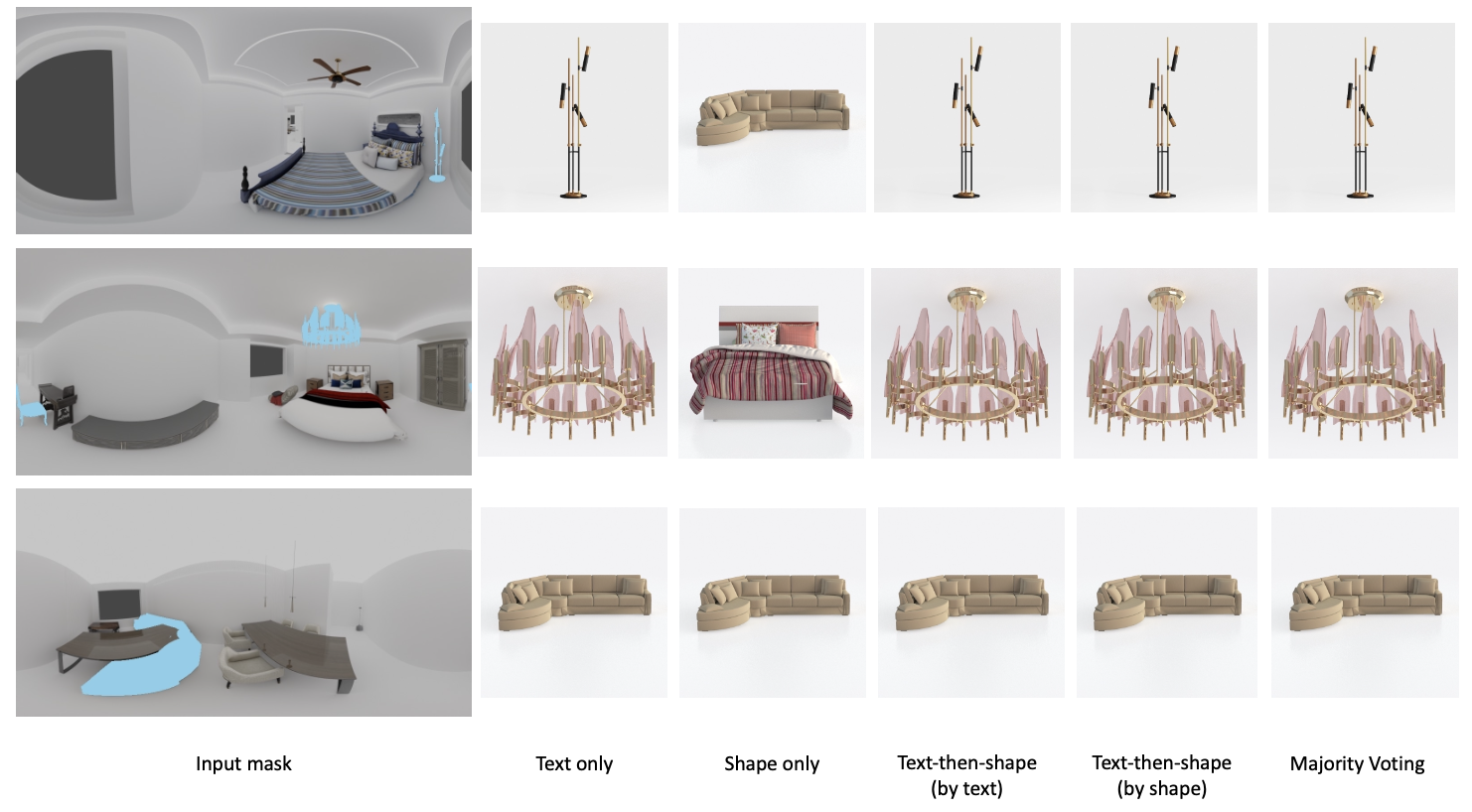}
    \caption{Qualitative comparison of retrieval strategies explored by team NoResources. From left to right: (1) the input scene with the masked query, (2) text-only retrieval, (3) shape-only retrieval, (4) text-then-shape (ranked by text), (5) text-then-shape (ranked by shape), and (6) the final result from the majority voting.}
    \label{fig:vis_NoResources}
\end{figure*}

Fig.~\ref{fig:vis_NoResources} compares the five retrieval heuristics explored by \emph{NoResources}. Text-only retrieval performs well when the query emphasizes linguistic detail (e.g.,  a multi-bulb floor lamp or an ornate chandelier) but struggles with shape-driven descriptions. Conversely, shape-only retrieval excels in structure-focused queries but ignores fine-grained semantic cues. The two-stage filtering by text and re-ranking by shape, and vice versa, resolves many of the errors made by single-modality methods. However, they may still diverge in final rankings. A simple weighted-vote ensemble resolves these conflicts by assigning greater weight to the staged strategies, effectively leveraging their complementary strengths. This ensemble consistently ranks the correct object among the top results. Collectively, the examples underscore that ROOMELSA’s key remaining challenge lies not in coarse recall, which is already saturated, but in fine-grained tie-breaking, best addressed through principled fusion of linguistic and geometric signals.

\section{Conclusion}\label{sec:conclusion}
In this work, we introduce the ROOMELSA challenge, establishing the first benchmark that directly links language grounding in 3D scenes to gallery-level CAD retrieval. In particular, the dataset comprises 44,445 mask-conditioned, attribute-rich queries paired with exhaustive mesh-level supervision and evaluates system performance using only rank-sensitive metrics. Results from 18 public submissions, with five achieving nearly perfect recall within the top ten candidates, demonstrate that modern language–shape encoders can effectively reduce a large-scale item gallery to a concise shortlist. However, these proposed models still encounter difficulties in ranking highly similar meshes. The top-performing pipelines reflect two complementary trends. First, multimodal fusion strategies that combine appearance, geometry, and spatial context help resolve most ranking ambiguities. Second, lightweight ensembles based on frozen CLIP embeddings remain competitive when enhanced with targeted pre-processing and simple voting schemes. These findings validate ROOMELSA as a benchmark for fine-grained 3D retrieval and offer insights for future work.

Two directions appear particularly promising for closing the remaining gap to perfect MRR. First, enriching the object gallery with material and part-level variants would require models to reason beyond coarse features such as silhouette and color. Second, deeper integration of spatial context, for example, through the prediction of object affordances or support relationships, could address the remaining rank-1 errors that result from functional mismatches rather than visual similarity.

\section*{CRediT Authorship Contribution Statement}


\textbf{Trong-Thuan Nguyen}: Conceptualization, Writing – Review $\&$ Editing, Project Administration, Supervision. 
\textbf{Viet-Tham Huynh}: Conceptualization, Writing – Review $\&$ Editing, Supervision. 
\textbf{Quang-Thuc Nguyen}: Conceptualization, Writing – Review $\&$ Editing, Supervision. 
\textbf{Hoang-Phuc Nguyen}: Data Curation, Software. 
\textbf{Long Le Bao}: Methodology, Writing – original draft. 
\textbf{Thai Hoang Minh}: Methodology, Writing – original draft. 
\textbf{Minh Nguyen Anh}: Methodology, Writing – original draft. 
\textbf{Thang Nguyen Tien}: Methodology, Writing – original draft. 
\textbf{Phat Nguyen Thuan}: Methodology, Writing – original draft. 
\textbf{Huy Nguyen Phong}: Methodology, Writing – original draft. 
\textbf{Bao Huynh Thai}: Methodology, Writing – original draft. 
\textbf{Vinh-Tiep Nguyen}: Methodology, Writing – original draft. 
\textbf{Duc-Vu Nguyen}: Methodology, Writing – original draft. 
\textbf{Phu-Hoa Pham}: Methodology, Writing – original draft. 
\textbf{Minh-Huy Le-Hoang}: Methodology, Writing – original draft. 
\textbf{Nguyen-Khang Le}: Methodology, Writing – original draft.
\textbf{Minh-Chinh Nguyen}: Methodology, Writing – original draft.
\textbf{Minh-Quan Ho}: Methodology, Writing – original draft.
\textbf{Ngoc-Long Tran}: Methodology, Writing – original draft.
\textbf{Hien-Long Le-Hoang}: Methodology, Writing – original draft.
\textbf{Man-Khoi Tran}: Methodology, Writing – original draft.
\textbf{Anh-Duong Tran}: Methodology, Writing – original draft.
\textbf{Kim Nguyen}: Methodology, Writing – original draft. 
\textbf{Quan Nguyen Hung}: Methodology, Writing – original draft. 
\textbf{Dat Phan Thanh}: Methodology, Writing – original draft. 
\textbf{Hoang Tran Van}: Methodology, Writing – original draft. 
\textbf{Tien Huynh Viet}: Methodology, Writing – original draft. 
\textbf{Nhan Nguyen Viet Thien}: Methodology, Writing – original draft. 
\textbf{Dinh-Khoi Vo}: Methodology, Writing – original draft. 
\textbf{Van-Loc Nguyen}: Methodology, Writing – original draft. 
\textbf{Trung-Nghia Le}: Methodology, Writing – original draft. 
\textbf{Tam V. Nguyen}: Conceptualization,  Supervision, Writing – Review $\&$ Editing. 
\textbf{Minh-Triet Tran}: Conceptualization, Supervision, Funding acquisition, Writing - Review \& Editing.

\section*{Data Availability}

\highlight{After the challenge concluded, the dataset has been made publicly available for academic purposes.}

\section*{Declaration of Competing Interest}

The authors declare that they have no known competing financial interests or personal relationships that could have appeared to influence the work reported in this paper.

\newpage    
\bibliographystyle{cag-num-names}
\bibliography{final}

\begin{thebibliography}{49}
\providecommand{\natexlab}[1]{#1}
\providecommand{\url}[1]{\texttt{#1}}
\providecommand{\href}[2]{#2}
\providecommand{\path}[1]{#1}
\providecommand{\eprint}[1]{\href{http://arxiv.org/abs/#1}{\path{#1}}}
\providecommand{\DOIprefix}{doi:}
\providecommand{\ArXivprefix}{arXiv:}
\providecommand{\URLprefix}{URL: }
\providecommand{\Pubmedprefix}{pmid:}
\providecommand{\doi}[1]{\href{http://dx.doi.org/#1}{\path{#1}}}
\providecommand{\Pubmed}[1]{\href{pmid:#1}{\path{#1}}}
\providecommand{\BIBand}{and}
\providecommand{\bibinfo}[2]{#2}
\ifx\xfnm\undefined \def\xfnm[#1]{\unskip,\space#1}\fi
\bibitem[{Dai et~al.(2017)Dai, Chang, Savva, Halber, Funkhouser and Nie{\ss}ner}]{dai2017scannet}
\bibinfo{author}{Dai\xfnm[ A]}, \bibinfo{author}{Chang\xfnm[ AX]}, \bibinfo{author}{Savva\xfnm[ M]}, \bibinfo{author}{Halber\xfnm[ M]}, \bibinfo{author}{Funkhouser\xfnm[ T]}, \bibinfo{author}{Nie{\ss}ner\xfnm[ M]}.
\newblock \bibinfo{title}{Scannet: Richly-annotated 3d reconstructions of indoor scenes}.
\newblock In: \bibinfo{booktitle}{Proceedings of the IEEE conference on computer vision and pattern recognition}. \bibinfo{year}{2017}, p. \bibinfo{pages}{5828--5839}.
\bibitem[{Chang et~al.(2017)Chang, Dai, Funkhouser, Halber, Niessner, Savva et~al.}]{chang2017matterport3d}
\bibinfo{author}{Chang\xfnm[ A]}, \bibinfo{author}{Dai\xfnm[ A]}, \bibinfo{author}{Funkhouser\xfnm[ T]}, \bibinfo{author}{Halber\xfnm[ M]}, \bibinfo{author}{Niessner\xfnm[ M]}, \bibinfo{author}{Savva\xfnm[ M]}, et~al.
\newblock \bibinfo{title}{Matterport3d: Learning from rgb-d data in indoor environments}.
\newblock \bibinfo{journal}{arXiv preprint arXiv:170906158} \bibinfo{year}{2017};.
\bibitem[{Fu et~al.(2021)Fu, Jia, Gao, Gong, Zhao, Maybank et~al.}]{fu20213d}
\bibinfo{author}{Fu\xfnm[ H]}, \bibinfo{author}{Jia\xfnm[ R]}, \bibinfo{author}{Gao\xfnm[ L]}, \bibinfo{author}{Gong\xfnm[ M]}, \bibinfo{author}{Zhao\xfnm[ B]}, \bibinfo{author}{Maybank\xfnm[ S]}, et~al.
\newblock \bibinfo{title}{3d-future: 3d furniture shape with texture}.
\newblock \bibinfo{journal}{International Journal of Computer Vision} \bibinfo{year}{2021};\bibinfo{volume}{129}:\bibinfo{pages}{3313--3337}.
\bibitem[{Achlioptas et~al.(2020)Achlioptas, Abdelreheem, Xia, Elhoseiny and Guibas}]{achlioptas2020referit3d}
\bibinfo{author}{Achlioptas\xfnm[ P]}, \bibinfo{author}{Abdelreheem\xfnm[ A]}, \bibinfo{author}{Xia\xfnm[ F]}, \bibinfo{author}{Elhoseiny\xfnm[ M]}, \bibinfo{author}{Guibas\xfnm[ L]}.
\newblock \bibinfo{title}{Referit3d: Neural listeners for fine-grained 3d object identification in real-world scenes}.
\newblock In: \bibinfo{booktitle}{Computer Vision--ECCV 2020: 16th European Conference, Glasgow, UK, August 23--28, 2020, Proceedings, Part I 16}. \bibinfo{organization}{Springer}; \bibinfo{year}{2020}, p. \bibinfo{pages}{422--440}.
\bibitem[{Chen et~al.(2020)Chen, Chang and Nie{\ss}ner}]{chen2020scanrefer}
\bibinfo{author}{Chen\xfnm[ DZ]}, \bibinfo{author}{Chang\xfnm[ AX]}, \bibinfo{author}{Nie{\ss}ner\xfnm[ M]}.
\newblock \bibinfo{title}{Scanrefer: 3d object localization in rgb-d scans using natural language}.
\newblock In: \bibinfo{booktitle}{European conference on computer vision}. \bibinfo{organization}{Springer}; \bibinfo{year}{2020}, p. \bibinfo{pages}{202--221}.
\bibitem[{Wang et~al.(2025)Wang, Gong and Chang}]{wang2025vigil3d}
\bibinfo{author}{Wang\xfnm[ AT]}, \bibinfo{author}{Gong\xfnm[ Z]}, \bibinfo{author}{Chang\xfnm[ AX]}.
\newblock \bibinfo{title}{Vigil3d: A linguistically diverse dataset for 3d visual grounding}.
\newblock \bibinfo{journal}{arXiv preprint arXiv:250101366} \bibinfo{year}{2025};.
\bibitem[{Abdul-Rashid et~al.(2018)Abdul-Rashid, Yuan, Li, Lu, Bai, Bai et~al.}]{Hameed-SHREC2018}
\bibinfo{author}{Abdul-Rashid\xfnm[ H]}, \bibinfo{author}{Yuan\xfnm[ J]}, \bibinfo{author}{Li\xfnm[ B]}, \bibinfo{author}{Lu\xfnm[ Y]}, \bibinfo{author}{Bai\xfnm[ S]}, \bibinfo{author}{Bai\xfnm[ X]}, et~al.
\newblock \bibinfo{title}{{2D Image-Based {3D} Scene Retrieval}}.
\newblock In: \bibinfo{editor}{Telea\xfnm[ A]}, \bibinfo{editor}{Theoharis\xfnm[ T]}, \bibinfo{editor}{Veltkamp\xfnm[ R]}, editors. \bibinfo{booktitle}{Eurographics Workshop on {3D} Object Retrieval}. \bibinfo{year}{2018},.
\bibitem[{Abdul-Rashid et~al.(2019)Abdul-Rashid, Yuan, Li, Lu, Schreck, Bui et~al.}]{Hameed-SHREC2019}
\bibinfo{author}{Abdul-Rashid\xfnm[ H]}, \bibinfo{author}{Yuan\xfnm[ J]}, \bibinfo{author}{Li\xfnm[ B]}, \bibinfo{author}{Lu\xfnm[ Y]}, \bibinfo{author}{Schreck\xfnm[ T]}, \bibinfo{author}{Bui\xfnm[ NM]}, et~al.
\newblock \bibinfo{title}{{SHREC}’19 track: Extended 2d scene image-based {3D} scene retrieval}.
\newblock \bibinfo{journal}{Eurographics Workshop on {3D} Object Retrieval} \bibinfo{year}{2019};\bibinfo{volume}{700}:\bibinfo{pages}{70}.
\bibitem[{Li et~al.(2019)Li, Liu, Nie, Song, Li, Wang et~al.}]{Li-SHREC2019}
\bibinfo{author}{Li\xfnm[ W]}, \bibinfo{author}{Liu\xfnm[ A]}, \bibinfo{author}{Nie\xfnm[ W]}, \bibinfo{author}{Song\xfnm[ D]}, \bibinfo{author}{Li\xfnm[ Y]}, \bibinfo{author}{Wang\xfnm[ W]}, et~al.
\newblock \bibinfo{title}{{SHREC} 2019-monocular image based {3D} model retrieval}.
\newblock In: \bibinfo{booktitle}{Eurographics Workshop {3D} Object Retrieval}. \bibinfo{year}{2019}, p. \bibinfo{pages}{1--8}.
\bibitem[{Li et~al.(2020)Li, Song, Liu, Nie, Zhang, Zhao et~al.}]{Li-SHREC2020}
\bibinfo{author}{Li\xfnm[ W]}, \bibinfo{author}{Song\xfnm[ D]}, \bibinfo{author}{Liu\xfnm[ A]}, \bibinfo{author}{Nie\xfnm[ W]}, \bibinfo{author}{Zhang\xfnm[ T]}, \bibinfo{author}{Zhao\xfnm[ X]}, et~al.
\newblock \bibinfo{title}{{SHREC} 2020 track: extended monocular image based 3d model retrieval}.
\newblock In: \bibinfo{booktitle}{Eurographics Workshop {3D} Object Retrieval}. \bibinfo{year}{2020},.
\bibitem[{Feng et~al.(2022)Feng, Gao, Zhao, Guo, Bagewadi, Bui et~al.}]{Feng-SHREC2022}
\bibinfo{author}{Feng\xfnm[ Y]}, \bibinfo{author}{Gao\xfnm[ Y]}, \bibinfo{author}{Zhao\xfnm[ X]}, \bibinfo{author}{Guo\xfnm[ Y]}, \bibinfo{author}{Bagewadi\xfnm[ N]}, \bibinfo{author}{Bui\xfnm[ NT]}, et~al.
\newblock \bibinfo{title}{{SHREC}’22 track: Open-set {3D} object retrieval}.
\newblock \bibinfo{journal}{Computers \& Graphics} \bibinfo{year}{2022};\bibinfo{volume}{107}:\bibinfo{pages}{231--240}.
\bibitem[{Qin et~al.(2022)Qin, Yuan, Chen, {Ben Amor}, Fang, Hoang-Xuan et~al.}]{Qin-SHREC2022}
\bibinfo{author}{Qin\xfnm[ J]}, \bibinfo{author}{Yuan\xfnm[ S]}, \bibinfo{author}{Chen\xfnm[ J]}, \bibinfo{author}{{Ben Amor}\xfnm[ B]}, \bibinfo{author}{Fang\xfnm[ Y]}, \bibinfo{author}{Hoang-Xuan\xfnm[ N]}, et~al.
\newblock \bibinfo{title}{Shrec’22 track: Sketch-based {3D} shape retrieval in the wild}.
\newblock \bibinfo{journal}{Computers and Graphics} \bibinfo{year}{2022};.
\bibitem[{Le et~al.(2023{\natexlab{a}})Le, Nguyen, Le, Nguyen, Huynh, Do et~al.}]{le2023sketchanimar}
\bibinfo{author}{Le\xfnm[ TN]}, \bibinfo{author}{Nguyen\xfnm[ TV]}, \bibinfo{author}{Le\xfnm[ MQ]}, \bibinfo{author}{Nguyen\xfnm[ TT]}, \bibinfo{author}{Huynh\xfnm[ VT]}, \bibinfo{author}{Do\xfnm[ TL]}, et~al.
\newblock \bibinfo{title}{Sketchanimar: Sketch-based 3d animal fine-grained retrieval}.
\newblock \bibinfo{journal}{Computers \& Graphics} \bibinfo{year}{2023}{\natexlab{a}};\bibinfo{volume}{116}:\bibinfo{pages}{150--161}.
\bibitem[{Le et~al.(2023{\natexlab{b}})Le, Nguyen, Le, Nguyen, Huynh, Do et~al.}]{le2023textanimar}
\bibinfo{author}{Le\xfnm[ TN]}, \bibinfo{author}{Nguyen\xfnm[ TV]}, \bibinfo{author}{Le\xfnm[ MQ]}, \bibinfo{author}{Nguyen\xfnm[ TT]}, \bibinfo{author}{Huynh\xfnm[ VT]}, \bibinfo{author}{Do\xfnm[ TL]}, et~al.
\newblock \bibinfo{title}{Textanimar: text-based 3d animal fine-grained retrieval}.
\newblock \bibinfo{journal}{Computers \& Graphics} \bibinfo{year}{2023}{\natexlab{b}};\bibinfo{volume}{116}:\bibinfo{pages}{162--172}.
\bibitem[{Zhang et~al.(2023)Zhang, Gong and Chang}]{zhang2023multi3drefer}
\bibinfo{author}{Zhang\xfnm[ Y]}, \bibinfo{author}{Gong\xfnm[ Z]}, \bibinfo{author}{Chang\xfnm[ AX]}.
\newblock \bibinfo{title}{Multi3drefer: Grounding text description to multiple 3d objects}.
\newblock In: \bibinfo{booktitle}{Proceedings of the IEEE/CVF International Conference on Computer Vision}. \bibinfo{year}{2023}, p. \bibinfo{pages}{15225--15236}.
\bibitem[{Kazemzadeh et~al.(2014)Kazemzadeh, Ordonez, Matten and Berg}]{kazemzadeh2014referitgame}
\bibinfo{author}{Kazemzadeh\xfnm[ S]}, \bibinfo{author}{Ordonez\xfnm[ V]}, \bibinfo{author}{Matten\xfnm[ M]}, \bibinfo{author}{Berg\xfnm[ T]}.
\newblock \bibinfo{title}{Referitgame: Referring to objects in photographs of natural scenes}.
\newblock In: \bibinfo{booktitle}{Proceedings of the 2014 conference on empirical methods in natural language processing (EMNLP)}. \bibinfo{year}{2014}, p. \bibinfo{pages}{787--798}.
\bibitem[{Mao et~al.(2016)Mao, Huang, Toshev, Camburu, Yuille and Murphy}]{mao2016generation}
\bibinfo{author}{Mao\xfnm[ J]}, \bibinfo{author}{Huang\xfnm[ J]}, \bibinfo{author}{Toshev\xfnm[ A]}, \bibinfo{author}{Camburu\xfnm[ O]}, \bibinfo{author}{Yuille\xfnm[ AL]}, \bibinfo{author}{Murphy\xfnm[ K]}.
\newblock \bibinfo{title}{Generation and comprehension of unambiguous object descriptions}.
\newblock In: \bibinfo{booktitle}{Proceedings of the IEEE conference on computer vision and pattern recognition}. \bibinfo{year}{2016}, p. \bibinfo{pages}{11--20}.
\bibitem[{Lin et~al.(2014)Lin, Maire, Belongie, Hays, Perona, Ramanan et~al.}]{lin2014microsoft}
\bibinfo{author}{Lin\xfnm[ TY]}, \bibinfo{author}{Maire\xfnm[ M]}, \bibinfo{author}{Belongie\xfnm[ S]}, \bibinfo{author}{Hays\xfnm[ J]}, \bibinfo{author}{Perona\xfnm[ P]}, \bibinfo{author}{Ramanan\xfnm[ D]}, et~al.
\newblock \bibinfo{title}{Microsoft coco: Common objects in context}.
\newblock In: \bibinfo{booktitle}{Computer vision--ECCV 2014: 13th European conference, zurich, Switzerland, September 6-12, 2014, proceedings, part v 13}. \bibinfo{organization}{Springer}; \bibinfo{year}{2014}, p. \bibinfo{pages}{740--755}.
\bibitem[{Yu et~al.(2018)Yu, Lin, Shen, Yang, Lu, Bansal et~al.}]{yu2018mattnet}
\bibinfo{author}{Yu\xfnm[ L]}, \bibinfo{author}{Lin\xfnm[ Z]}, \bibinfo{author}{Shen\xfnm[ X]}, \bibinfo{author}{Yang\xfnm[ J]}, \bibinfo{author}{Lu\xfnm[ X]}, \bibinfo{author}{Bansal\xfnm[ M]}, et~al.
\newblock \bibinfo{title}{Mattnet: Modular attention network for referring expression comprehension}.
\newblock In: \bibinfo{booktitle}{Proceedings of the IEEE conference on computer vision and pattern recognition}. \bibinfo{year}{2018}, p. \bibinfo{pages}{1307--1315}.
\bibitem[{Yang et~al.(2022)Yang, Wang, Tang, Chen, Zhao and Torr}]{yang2022lavt}
\bibinfo{author}{Yang\xfnm[ Z]}, \bibinfo{author}{Wang\xfnm[ J]}, \bibinfo{author}{Tang\xfnm[ Y]}, \bibinfo{author}{Chen\xfnm[ K]}, \bibinfo{author}{Zhao\xfnm[ H]}, \bibinfo{author}{Torr\xfnm[ PH]}.
\newblock \bibinfo{title}{Lavt: Language-aware vision transformer for referring image segmentation}.
\newblock In: \bibinfo{booktitle}{Proceedings of the IEEE/CVF conference on computer vision and pattern recognition}. \bibinfo{year}{2022}, p. \bibinfo{pages}{18155--18165}.
\bibitem[{Wu et~al.(2022)Wu, Jiang, Sun, Yuan and Luo}]{wu2022language}
\bibinfo{author}{Wu\xfnm[ J]}, \bibinfo{author}{Jiang\xfnm[ Y]}, \bibinfo{author}{Sun\xfnm[ P]}, \bibinfo{author}{Yuan\xfnm[ Z]}, \bibinfo{author}{Luo\xfnm[ P]}.
\newblock \bibinfo{title}{Language as queries for referring video object segmentation}.
\newblock In: \bibinfo{booktitle}{Proceedings of the IEEE/CVF Conference on Computer Vision and Pattern Recognition}. \bibinfo{year}{2022}, p. \bibinfo{pages}{4974--4984}.
\bibitem[{Cheng et~al.(2022)Cheng, Misra, Schwing, Kirillov and Girdhar}]{cheng2022masked}
\bibinfo{author}{Cheng\xfnm[ B]}, \bibinfo{author}{Misra\xfnm[ I]}, \bibinfo{author}{Schwing\xfnm[ AG]}, \bibinfo{author}{Kirillov\xfnm[ A]}, \bibinfo{author}{Girdhar\xfnm[ R]}.
\newblock \bibinfo{title}{Masked-attention mask transformer for universal image segmentation}.
\newblock In: \bibinfo{booktitle}{Proceedings of the IEEE/CVF conference on computer vision and pattern recognition}. \bibinfo{year}{2022}, p. \bibinfo{pages}{1290--1299}.
\bibitem[{Mo et~al.(2019)Mo, Zhu, Chang, Yi, Tripathi, Guibas et~al.}]{mo2019partnet}
\bibinfo{author}{Mo\xfnm[ K]}, \bibinfo{author}{Zhu\xfnm[ S]}, \bibinfo{author}{Chang\xfnm[ AX]}, \bibinfo{author}{Yi\xfnm[ L]}, \bibinfo{author}{Tripathi\xfnm[ S]}, \bibinfo{author}{Guibas\xfnm[ LJ]}, et~al.
\newblock \bibinfo{title}{Partnet: A large-scale benchmark for fine-grained and hierarchical part-level 3d object understanding}.
\newblock In: \bibinfo{booktitle}{Proceedings of the IEEE/CVF conference on computer vision and pattern recognition}. \bibinfo{year}{2019}, p. \bibinfo{pages}{909--918}.
\bibitem[{Liu et~al.(2024)Liu, Obukhov, Wegner and Schindler}]{liu2024point2cad}
\bibinfo{author}{Liu\xfnm[ Y]}, \bibinfo{author}{Obukhov\xfnm[ A]}, \bibinfo{author}{Wegner\xfnm[ JD]}, \bibinfo{author}{Schindler\xfnm[ K]}.
\newblock \bibinfo{title}{Point2cad: Reverse engineering cad models from 3d point clouds}.
\newblock In: \bibinfo{booktitle}{Proceedings of the IEEE/CVF Conference on Computer Vision and Pattern Recognition}. \bibinfo{year}{2024}, p. \bibinfo{pages}{3763--3772}.
\bibitem[{Schult et~al.(2023)Schult, Engelmann, Hermans, Litany, Tang and Leibe}]{schult2023mask3d}
\bibinfo{author}{Schult\xfnm[ J]}, \bibinfo{author}{Engelmann\xfnm[ F]}, \bibinfo{author}{Hermans\xfnm[ A]}, \bibinfo{author}{Litany\xfnm[ O]}, \bibinfo{author}{Tang\xfnm[ S]}, \bibinfo{author}{Leibe\xfnm[ B]}.
\newblock \bibinfo{title}{Mask3d: Mask transformer for 3d semantic instance segmentation}.
\newblock In: \bibinfo{booktitle}{2023 IEEE International Conference on Robotics and Automation (ICRA)}. \bibinfo{organization}{IEEE}; \bibinfo{year}{2023}, p. \bibinfo{pages}{8216--8223}.
\bibitem[{He and Ding(2024)}]{he2024refmask3d}
\bibinfo{author}{He\xfnm[ S]}, \bibinfo{author}{Ding\xfnm[ H]}.
\newblock \bibinfo{title}{Refmask3d: Language-guided transformer for 3d referring segmentation}.
\newblock In: \bibinfo{booktitle}{Proceedings of the 32nd ACM International Conference on Multimedia}. \bibinfo{year}{2024}, p. \bibinfo{pages}{8316--8325}.
\bibitem[{Lee et~al.(2024)Lee, Zhao and Lee}]{lee2024segment}
\bibinfo{author}{Lee\xfnm[ S]}, \bibinfo{author}{Zhao\xfnm[ Y]}, \bibinfo{author}{Lee\xfnm[ GH]}.
\newblock \bibinfo{title}{Segment any 3d object with language}.
\newblock \bibinfo{journal}{arXiv preprint arXiv:240402157} \bibinfo{year}{2024};.
\bibitem[{Radford et~al.(2021)Radford, Kim, Hallacy, Ramesh, Goh, Agarwal et~al.}]{radford2021learning}
\bibinfo{author}{Radford\xfnm[ A]}, \bibinfo{author}{Kim\xfnm[ JW]}, \bibinfo{author}{Hallacy\xfnm[ C]}, \bibinfo{author}{Ramesh\xfnm[ A]}, \bibinfo{author}{Goh\xfnm[ G]}, \bibinfo{author}{Agarwal\xfnm[ S]}, et~al.
\newblock \bibinfo{title}{Learning transferable visual models from natural language supervision}.
\newblock In: \bibinfo{booktitle}{International conference on machine learning}. \bibinfo{organization}{PmLR}; \bibinfo{year}{2021}, p. \bibinfo{pages}{8748--8763}.
\bibitem[{Xue et~al.(2023)Xue, Gao, Xing, Mart{\'\i}n-Mart{\'\i}n, Wu, Xiong et~al.}]{xue2023ulip}
\bibinfo{author}{Xue\xfnm[ L]}, \bibinfo{author}{Gao\xfnm[ M]}, \bibinfo{author}{Xing\xfnm[ C]}, \bibinfo{author}{Mart{\'\i}n-Mart{\'\i}n\xfnm[ R]}, \bibinfo{author}{Wu\xfnm[ J]}, \bibinfo{author}{Xiong\xfnm[ C]}, et~al.
\newblock \bibinfo{title}{Ulip: Learning a unified representation of language, images, and point clouds for 3d understanding}.
\newblock In: \bibinfo{booktitle}{Proceedings of the IEEE/CVF conference on computer vision and pattern recognition}. \bibinfo{year}{2023}, p. \bibinfo{pages}{1179--1189}.
\bibitem[{Xue et~al.(2024)Xue, Yu, Zhang, Panagopoulou, Li, Mart{\'\i}n-Mart{\'\i}n et~al.}]{xue2024ulip}
\bibinfo{author}{Xue\xfnm[ L]}, \bibinfo{author}{Yu\xfnm[ N]}, \bibinfo{author}{Zhang\xfnm[ S]}, \bibinfo{author}{Panagopoulou\xfnm[ A]}, \bibinfo{author}{Li\xfnm[ J]}, \bibinfo{author}{Mart{\'\i}n-Mart{\'\i}n\xfnm[ R]}, et~al.
\newblock \bibinfo{title}{Ulip-2: Towards scalable multimodal pre-training for 3d understanding}.
\newblock In: \bibinfo{booktitle}{Proceedings of the IEEE/CVF Conference on Computer Vision and Pattern Recognition}. \bibinfo{year}{2024}, p. \bibinfo{pages}{27091--27101}.
\bibitem[{Deitke et~al.(2023)Deitke, Liu, Wallingford, Ngo, Michel, Kusupati et~al.}]{deitke2023objaverse}
\bibinfo{author}{Deitke\xfnm[ M]}, \bibinfo{author}{Liu\xfnm[ R]}, \bibinfo{author}{Wallingford\xfnm[ M]}, \bibinfo{author}{Ngo\xfnm[ H]}, \bibinfo{author}{Michel\xfnm[ O]}, \bibinfo{author}{Kusupati\xfnm[ A]}, et~al.
\newblock \bibinfo{title}{Objaverse-xl: A universe of 10m+ 3d objects}.
\newblock \bibinfo{journal}{Advances in Neural Information Processing Systems} \bibinfo{year}{2023};\bibinfo{volume}{36}:\bibinfo{pages}{35799--35813}.
\bibitem[{Xiong et~al.(2025)Xiong, Zhuge, Zhu, Zhang and Lu}]{xiong20253ur}
\bibinfo{author}{Xiong\xfnm[ H]}, \bibinfo{author}{Zhuge\xfnm[ Y]}, \bibinfo{author}{Zhu\xfnm[ J]}, \bibinfo{author}{Zhang\xfnm[ L]}, \bibinfo{author}{Lu\xfnm[ H]}.
\newblock \bibinfo{title}{3ur-llm: An end-to-end multimodal large language model for 3d scene understanding}.
\newblock \bibinfo{journal}{arXiv preprint arXiv:250107819} \bibinfo{year}{2025};.
\bibitem[{Li et~al.(2014)Li, Lu, Li, Godil, Schreck, Aono et~al.}]{Li-SHREC2014}
\bibinfo{author}{Li\xfnm[ B]}, \bibinfo{author}{Lu\xfnm[ Y]}, \bibinfo{author}{Li\xfnm[ C]}, \bibinfo{author}{Godil\xfnm[ A]}, \bibinfo{author}{Schreck\xfnm[ T]}, \bibinfo{author}{Aono\xfnm[ M]}, et~al.
\newblock \bibinfo{title}{Shrec’14 track: Extended large scale sketch-based {3D} shape retrieval}.
\newblock In: \bibinfo{booktitle}{Eurographics workshop on {3D} object retrieval}; vol. \bibinfo{volume}{2014}. \bibinfo{year}{2014}, p. \bibinfo{pages}{121--130}.
\bibitem[{Yuan et~al.(2019)Yuan, Abdul-Rashid, Li, Lu, Schreck, Bui et~al.}]{Juefei-SHREC2019}
\bibinfo{author}{Yuan\xfnm[ J]}, \bibinfo{author}{Abdul-Rashid\xfnm[ H]}, \bibinfo{author}{Li\xfnm[ B]}, \bibinfo{author}{Lu\xfnm[ Y]}, \bibinfo{author}{Schreck\xfnm[ T]}, \bibinfo{author}{Bui\xfnm[ NM]}, et~al.
\newblock \bibinfo{title}{Shrec’19 track: Extended 2d scene sketch-based {3D} scene retrieval}.
\newblock \bibinfo{journal}{Eurographics Workshop on {3D} Object Retrieval} \bibinfo{year}{2019};\bibinfo{volume}{18}:\bibinfo{pages}{70}.
\bibitem[{Touvron et~al.(2023)Touvron, Lavril, Izacard, Martinet, Lachaux, Lacroix et~al.}]{touvron2023llama}
\bibinfo{author}{Touvron\xfnm[ H]}, \bibinfo{author}{Lavril\xfnm[ T]}, \bibinfo{author}{Izacard\xfnm[ G]}, \bibinfo{author}{Martinet\xfnm[ X]}, \bibinfo{author}{Lachaux\xfnm[ MA]}, \bibinfo{author}{Lacroix\xfnm[ T]}, et~al.
\newblock \bibinfo{title}{Llama: Open and efficient foundation language models}.
\newblock \bibinfo{journal}{arXiv preprint arXiv:230213971} \bibinfo{year}{2023};.
\bibitem[{Zhai et~al.(2023)Zhai, Mustafa, Kolesnikov and Beyer}]{zhai2023sigmoid}
\bibinfo{author}{Zhai\xfnm[ X]}, \bibinfo{author}{Mustafa\xfnm[ B]}, \bibinfo{author}{Kolesnikov\xfnm[ A]}, \bibinfo{author}{Beyer\xfnm[ L]}.
\newblock \bibinfo{title}{Sigmoid loss for language image pre-training}.
\newblock In: \bibinfo{booktitle}{Proceedings of the IEEE/CVF international conference on computer vision}. \bibinfo{year}{2023}, p. \bibinfo{pages}{11975--11986}.
\bibitem[{Chung et~al.(2024)Chung, Hou, Longpre, Zoph, Tay, Fedus et~al.}]{chung2024scaling}
\bibinfo{author}{Chung\xfnm[ HW]}, \bibinfo{author}{Hou\xfnm[ L]}, \bibinfo{author}{Longpre\xfnm[ S]}, \bibinfo{author}{Zoph\xfnm[ B]}, \bibinfo{author}{Tay\xfnm[ Y]}, \bibinfo{author}{Fedus\xfnm[ W]}, et~al.
\newblock \bibinfo{title}{Scaling instruction-finetuned language models}.
\newblock \bibinfo{journal}{Journal of Machine Learning Research} \bibinfo{year}{2024};\bibinfo{volume}{25}(\bibinfo{number}{70}):\bibinfo{pages}{1--53}.
\bibitem[{Xiao et~al.(2024{\natexlab{a}})Xiao, Liu, Zhang, Muennighoff, Lian and Nie}]{xiao2024c}
\bibinfo{author}{Xiao\xfnm[ S]}, \bibinfo{author}{Liu\xfnm[ Z]}, \bibinfo{author}{Zhang\xfnm[ P]}, \bibinfo{author}{Muennighoff\xfnm[ N]}, \bibinfo{author}{Lian\xfnm[ D]}, \bibinfo{author}{Nie\xfnm[ JY]}.
\newblock \bibinfo{title}{C-pack: Packed resources for general chinese embeddings}.
\newblock In: \bibinfo{booktitle}{Proceedings of the 47th international ACM SIGIR conference on research and development in information retrieval}. \bibinfo{year}{2024}{\natexlab{a}}, p. \bibinfo{pages}{641--649}.
\bibitem[{Li et~al.(2023)Li, Li, Savarese and Hoi}]{li2023blip}
\bibinfo{author}{Li\xfnm[ J]}, \bibinfo{author}{Li\xfnm[ D]}, \bibinfo{author}{Savarese\xfnm[ S]}, \bibinfo{author}{Hoi\xfnm[ S]}.
\newblock \bibinfo{title}{Blip-2: Bootstrapping language-image pre-training with frozen image encoders and large language models}.
\newblock In: \bibinfo{booktitle}{International conference on machine learning}. \bibinfo{organization}{PMLR}; \bibinfo{year}{2023}, p. \bibinfo{pages}{19730--19742}.
\bibitem[{Yu et~al.(2022)Yu, Tang, Rao, Huang, Zhou and Lu}]{yu2022point}
\bibinfo{author}{Yu\xfnm[ X]}, \bibinfo{author}{Tang\xfnm[ L]}, \bibinfo{author}{Rao\xfnm[ Y]}, \bibinfo{author}{Huang\xfnm[ T]}, \bibinfo{author}{Zhou\xfnm[ J]}, \bibinfo{author}{Lu\xfnm[ J]}.
\newblock \bibinfo{title}{Point-bert: Pre-training 3d point cloud transformers with masked point modeling}.
\newblock In: \bibinfo{booktitle}{Proceedings of the IEEE/CVF conference on computer vision and pattern recognition}. \bibinfo{year}{2022}, p. \bibinfo{pages}{19313--19322}.
\bibitem[{Liu et~al.(2023)Liu, Shi, Kuang, Zhu, Li, Han et~al.}]{liu2023openshape}
\bibinfo{author}{Liu\xfnm[ M]}, \bibinfo{author}{Shi\xfnm[ R]}, \bibinfo{author}{Kuang\xfnm[ K]}, \bibinfo{author}{Zhu\xfnm[ Y]}, \bibinfo{author}{Li\xfnm[ X]}, \bibinfo{author}{Han\xfnm[ S]}, et~al.
\newblock \bibinfo{title}{Openshape: Scaling up 3d shape representation towards open-world understanding}.
\newblock \bibinfo{journal}{Advances in neural information processing systems} \bibinfo{year}{2023};\bibinfo{volume}{36}:\bibinfo{pages}{44860--44879}.
\bibitem[{Shen et~al.(2022)Shen, Lin, Liao, Nie, Zheng and Zhao}]{shen2022panoformer}
\bibinfo{author}{Shen\xfnm[ Z]}, \bibinfo{author}{Lin\xfnm[ C]}, \bibinfo{author}{Liao\xfnm[ K]}, \bibinfo{author}{Nie\xfnm[ L]}, \bibinfo{author}{Zheng\xfnm[ Z]}, \bibinfo{author}{Zhao\xfnm[ Y]}.
\newblock \bibinfo{title}{Panoformer: panorama transformer for indoor $360^\circ$ depth estimation}.
\newblock In: \bibinfo{booktitle}{European Conference on Computer Vision}. \bibinfo{organization}{Springer}; \bibinfo{year}{2022}, p. \bibinfo{pages}{195--211}.
\bibitem[{Xiao et~al.(2024{\natexlab{b}})Xiao, Wu, Xu, Dai, Hu, Lu et~al.}]{xiao2024florence}
\bibinfo{author}{Xiao\xfnm[ B]}, \bibinfo{author}{Wu\xfnm[ H]}, \bibinfo{author}{Xu\xfnm[ W]}, \bibinfo{author}{Dai\xfnm[ X]}, \bibinfo{author}{Hu\xfnm[ H]}, \bibinfo{author}{Lu\xfnm[ Y]}, et~al.
\newblock \bibinfo{title}{Florence-2: Advancing a unified representation for a variety of vision tasks}.
\newblock In: \bibinfo{booktitle}{Proceedings of the IEEE/CVF Conference on Computer Vision and Pattern Recognition}. \bibinfo{year}{2024}{\natexlab{b}}, p. \bibinfo{pages}{4818--4829}.
\bibitem[{Khanam and Hussain(2024)}]{khanam2024yolov11}
\bibinfo{author}{Khanam\xfnm[ R]}, \bibinfo{author}{Hussain\xfnm[ M]}.
\newblock \bibinfo{title}{Yolov11: An overview of the key architectural enhancements}.
\newblock \bibinfo{journal}{arXiv preprint arXiv:241017725} \bibinfo{year}{2024};.
\bibitem[{An et~al.(2024)An, Bae, Choi, Jungkyu~Choi, Choi, Hong et~al.}]{an2024exaone}
\bibinfo{author}{An\xfnm[ S]}, \bibinfo{author}{Bae\xfnm[ K]}, \bibinfo{author}{Choi\xfnm[ E]}, \bibinfo{author}{Jungkyu~Choi\xfnm[ S]}, \bibinfo{author}{Choi\xfnm[ Y]}, \bibinfo{author}{Hong\xfnm[ S]}, et~al.
\newblock \bibinfo{title}{Exaone 3.0 7.8 b instruction tuned language model}.
\newblock \bibinfo{journal}{arXiv e-prints} \bibinfo{year}{2024};:\bibinfo{pages}{arXiv--2408}.
\bibitem[{Ohanyan et~al.(2024)Ohanyan, Manukyan, Wang, Navasardyan and Shi}]{ohanyan2024zero}
\bibinfo{author}{Ohanyan\xfnm[ M]}, \bibinfo{author}{Manukyan\xfnm[ H]}, \bibinfo{author}{Wang\xfnm[ Z]}, \bibinfo{author}{Navasardyan\xfnm[ S]}, \bibinfo{author}{Shi\xfnm[ H]}.
\newblock \bibinfo{title}{Zero-painter: Training-free layout control for text-to-image synthesis}.
\newblock In: \bibinfo{booktitle}{Proceedings of the IEEE/CVF Conference on Computer Vision and Pattern Recognition}. \bibinfo{year}{2024}, p. \bibinfo{pages}{8764--8774}.
\bibitem[{Wang et~al.(2024)Wang, Chen, Wang, Cao, Liu, Gao et~al.}]{wang2024enhancing}
\bibinfo{author}{Wang\xfnm[ W]}, \bibinfo{author}{Chen\xfnm[ Z]}, \bibinfo{author}{Wang\xfnm[ W]}, \bibinfo{author}{Cao\xfnm[ Y]}, \bibinfo{author}{Liu\xfnm[ Y]}, \bibinfo{author}{Gao\xfnm[ Z]}, et~al.
\newblock \bibinfo{title}{Enhancing the reasoning ability of multimodal large language models via mixed preference optimization}.
\newblock \bibinfo{journal}{arXiv preprint arXiv:241110442} \bibinfo{year}{2024};.
\bibitem[{Cherti et~al.(2023)Cherti, Beaumont, Wightman, Wortsman, Ilharco, Gordon et~al.}]{cherti2023reproducible}
\bibinfo{author}{Cherti\xfnm[ M]}, \bibinfo{author}{Beaumont\xfnm[ R]}, \bibinfo{author}{Wightman\xfnm[ R]}, \bibinfo{author}{Wortsman\xfnm[ M]}, \bibinfo{author}{Ilharco\xfnm[ G]}, \bibinfo{author}{Gordon\xfnm[ C]}, et~al.
\newblock \bibinfo{title}{Reproducible scaling laws for contrastive language-image learning}.
\newblock In: \bibinfo{booktitle}{Proceedings of the IEEE/CVF conference on computer vision and pattern recognition}. \bibinfo{year}{2023}, p. \bibinfo{pages}{2818--2829}.
\bibitem[{Chen et~al.(2024)Chen, Wang, Cao, Liu, Gao, Cui et~al.}]{chen2024expanding}
\bibinfo{author}{Chen\xfnm[ Z]}, \bibinfo{author}{Wang\xfnm[ W]}, \bibinfo{author}{Cao\xfnm[ Y]}, \bibinfo{author}{Liu\xfnm[ Y]}, \bibinfo{author}{Gao\xfnm[ Z]}, \bibinfo{author}{Cui\xfnm[ E]}, et~al.
\newblock \bibinfo{title}{Expanding performance boundaries of open-source multimodal models with model, data, and test-time scaling}.
\newblock \bibinfo{journal}{arXiv preprint arXiv:241205271} \bibinfo{year}{2024};.

\end{thebibliography}

\end{document}